\documentclass{article}
\usepackage{iclr2027_conference,times}

\usepackage{amsmath,amsfonts,bm}

\def\eqref#1{equation~\ref{#1}}

\def\1{\bm{1}}

\DeclareMathAlphabet{\mathsfit}{\encodingdefault}{\sfdefault}{m}{sl}
\SetMathAlphabet{\mathsfit}{bold}{\encodingdefault}{\sfdefault}{bx}{n}

\usepackage{hyperref}
\usepackage{url}
\usepackage{graphicx}
\usepackage{multicol}
\usepackage{multirow}
\usepackage{caption}

\title{The Devil Is in the Reconstruction Loss Scale: Rethinking Optimization in LLM Quantization}

\author{%
    Chao Li, Shigeng Wang, Anbang Yao\thanks{\ Corresponding author.}\\
  Intel Labs China\\
  \texttt{\{chao3.li,shigeng.wang,anbang.yao\}@intel.com} \\
}

\iclrfinalcopy
\begin{document}

\maketitle
\fancyhead{}

\begin{abstract} 
Post-training quantization (PTQ) methods typically quantize a pre-trained large language model (LLM) sequentially using a small set of calibration samples, achieving high data and computational efficiency. Specifically, the model is partitioned into a series of units, such as transformer blocks, with one unit quantized at each stage. State-of-the-art PTQ methods are predominantly learning-based, optimizing auxiliary quantization parameters (e.g., scaling factors, rotation matrices, clipping thresholds, and adapters) via gradient descent to minimize a reconstruction loss, thereby mitigating the impact of outliers and reducing quantization errors. In learning-based PTQ, a common practice is to use mean squared error (MSE) as the reconstruction loss function, yet its induced optimization behavior remains largely unexplored. In this work, we take a holistic view of sequential quantization and systematically investigate how optimization evolves from the first quantization stage to the last, aiming for a deep understanding of optimization in learning-based PTQ schemes. Through extensive empirical studies spanning representative learning-based PTQ methods, model families, model scales, architectures, and quantization settings, we consistently uncover \textit{Optimization Imbalance}: reconstruction loss magnitudes vary dramatically across stages, accompanied by highly uneven gradient magnitudes and parameter updates under MSE. We term the cross-stage range of loss magnitudes the \textit{reconstruction loss scale}. Our analysis reveals that MSE translates the unexpectedly large reconstruction loss scale into highly uneven gradient magnitudes, which in turn lead to uneven optimization strength across quantization stages. This finding suggests a general principle for improving learning-based PTQ: optimization strength across stages should be decoupled from the reconstruction loss scale. Theoretically, we show that root mean squared error (RMSE) naturally realizes this principle through implicit gradient normalization, thereby enabling balanced optimization across quantization stages. Moreover, RMSE can be naturally defined over different feature groupings, yielding four variants at the sample, channel, token, and element levels. As a drop-in replacement for MSE, RMSE consistently improves quantization performance across language modeling, commonsense reasoning, mathematical reasoning, and coding tasks, with all four variants outperforming MSE and the gains becoming increasingly pronounced under more aggressive quantization. Code will be available at \url{https://github.com/IntelChina-AI/RMSE}.
\end{abstract}

\begin{figure*}[!ht]
    \vskip -0.1 in
    \centering
    \centerline{\includegraphics[width=0.99\textwidth]{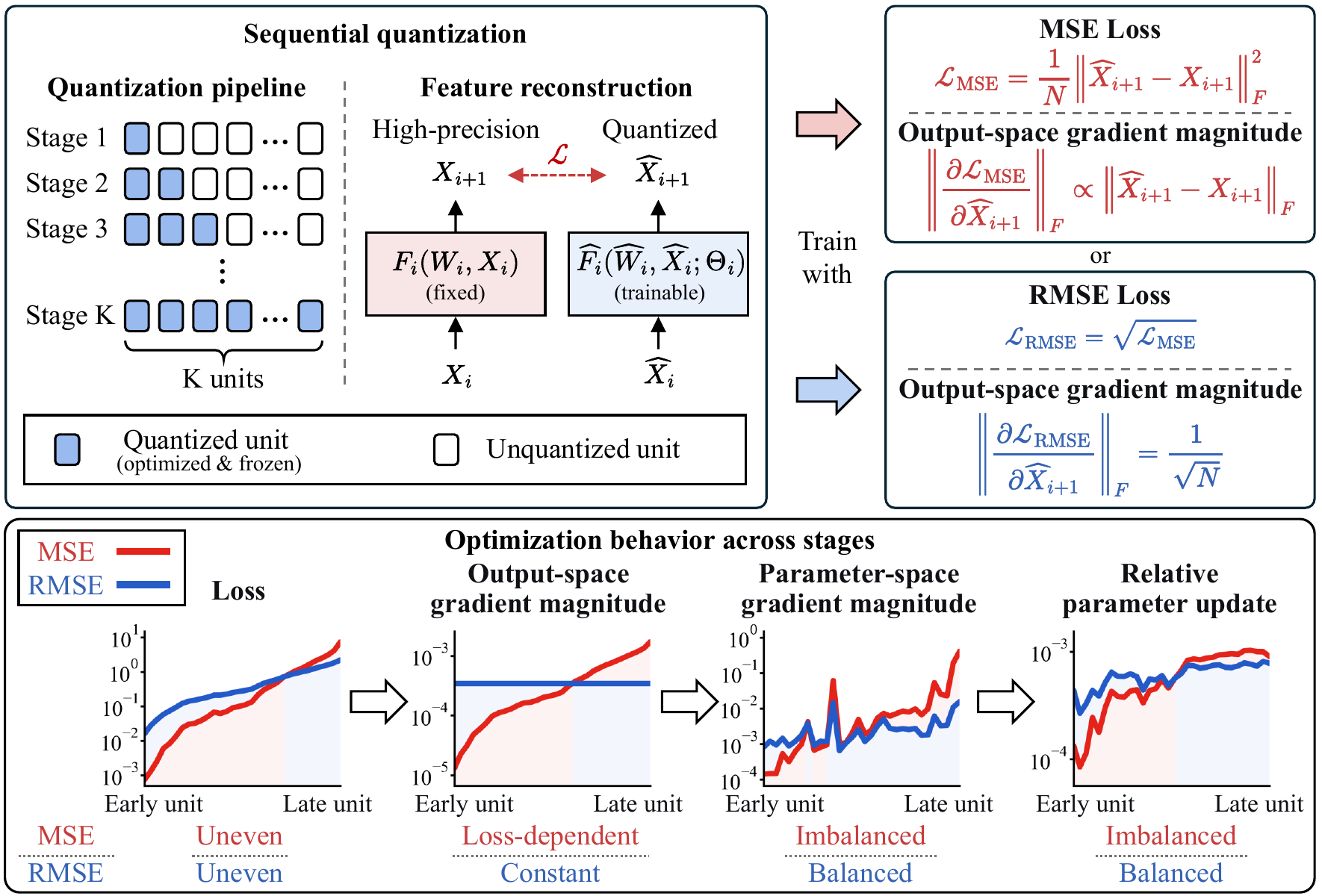}}
    \vskip -0.07 in
    \caption{\textbf{Illustration of Optimization Imbalance in learning-based PTQ and its mitigation via RMSE.} Learning-based PTQ methods predominantly use MSE as the reconstruction loss function for each stage in sequential quantization. By tracking optimization dynamics from the first stage to the last, we observe that reconstruction loss magnitudes vary dramatically across quantization stages. Our analysis reveals that MSE translates these differences into highly uneven gradients and parameter updates, giving rise to Optimization Imbalance. RMSE breaks this loss-dependent coupling through implicit gradient normalization, enabling more balanced optimization across quantization stages. Statistics are collected using OmniQuant on Qwen3.5-9B under W4A4 quantization.}
    \label{fig:method_main}
    \vskip -0.1 in
\end{figure*}

\section{Introduction}

Large language models (LLMs)~\citep{transformer, gpt3, gpt4, gemini, llama2, llama3, deepseekv3, deepseek-r1, qwen3.5} have achieved remarkable success across language understanding, generation, and reasoning tasks, yet their growing memory and computational demands pose a major obstacle to practical deployment. Post-training quantization (PTQ) offers an efficient solution by converting pre-trained high-precision models into low-bit replacements using a small calibration set, without the costly full-parameter retraining required by quantization-aware training~\mbox{\citep{q8bert,bitnet-1.58,onebit}}. To efficiently quantize LLMs, PTQ methods generally partition the model into a sequence of units (e.g., transformer blocks) and quantize them 
in a stage-wise manner, processing one unit at each stage while keeping previously quantized units fixed. We refer to this popular stage-wise PTQ paradigm as \textit{sequential quantization}. 

Under sequential quantization, existing PTQ methods differ primarily in how they perform feature (i.e., output) reconstruction to quantize each unit. Many methods employ carefully designed analytical or statistics-driven schemes for effective reconstruction, including second-order error compensation, activation-aware scaling, equivalent transformations, and fixed rotations. Representative methods include GPTQ~\citep{gptq}, AWQ~\citep{awq}, SmoothQuant~\citep{smoothquant}, QuaRot~\citep{quarot}, and others~\citep{gptaq, perq, wush, osaq}. By contrast, learning-based PTQ brings gradient descent into the feature reconstruction process without full-parameter retraining. At each quantization stage, the pre-trained weights remain frozen, while only a small set of auxiliary quantization parameters is optimized through feature reconstruction. This learning-based concept is adopted by most top-performing PTQ methods, which primarily differ in their learnable parameterizations and reconstruction schemes. OmniQuant~\citep{omniquant} learns scaling factors and clipping thresholds, DuQuant~\citep{duquant} optimizes orthogonal rotations and permutations, FlatQuant~\citep{flatquant} learns affine transformations, and SliderQuant~\citep{sliderquant} combines scaling and low-rank adaptation with sliding-window reconstruction. Other recent methods further explore related learnable design choices~\citep{spinquant, cbq, fptquant, paroquant}. Despite differences in their formulations, we observe that nearly all learning-based PTQ methods use mean squared error (MSE) as the reconstruction loss function, with the exception of LRQuant~\citep{lrquant}, which augments MSE with a cosine-similarity-based term in its proposed learnable smoothing strategy to address outliers. Although learning-based PTQ has advanced considerably, the optimization behavior induced by MSE throughout sequential quantization remains largely unexplored. Additionally, sequential quantization implicitly assumes that every unit receives sufficient optimization, yet this assumption has not been systematically examined. 

Motivated by these two largely overlooked perspectives, we take a holistic view of the sequential learning-based PTQ process, tracking how reconstruction losses, gradients, and parameter updates evolve from the first quantization stage to the last. Our empirical studies cover representative learning-based PTQ methods (OmniQuant, DuQuant, FlatQuant, and SliderQuant), model families (LLaMA2, LLaMA3.1, and Qwen3.5), model scales (7B, 8B, 9B, 27B, and 35B-A3B), architectures (dense and MoE), and quantization settings (W4A16, W3A16, W2A16, W2A16G128, and W4A4). Across this broad testing spectrum, we consistently uncover \textit{Optimization Imbalance}, as illustrated in Fig.~\ref{fig:method_main}: reconstruction loss magnitudes span an unexpectedly large cross-stage range, which we term the \emph{reconstruction loss scale}, with the largest often exceeding the smallest by more than $10{,}000\times$. Under MSE, this broad loss scale is accompanied by highly uneven gradients and parameter updates across quantization stages. 
This imbalance is consequential in sequential quantization because quantization errors propagate and accumulate across stages, meaning that insufficient optimization at one stage can negatively affect the quantization of all subsequent units.

Our analysis identifies MSE as the fundamental source of Optimization Imbalance, revealing that it inherently couples optimization strength with reconstruction loss magnitude. Specifically, MSE translates the unexpectedly large reconstruction loss scale into large differences in output-space gradient magnitude, which are further reflected in parameter-space gradients and ultimately manifest as highly uneven parameter updates across quantization stages. This loss-dependent coupling directly explains the observed Optimization Imbalance and suggests a general principle for improving learning-based PTQ: optimization strength across quantization stages should be decoupled from the reconstruction loss scale. This principle naturally leads to root mean squared error (RMSE). As a square-root transformation of MSE, RMSE preserves the gradient direction and relative magnitudes among gradient elements while implicitly normalizing the overall output-space gradient magnitude to a constant. RMSE therefore breaks the loss-dependent coupling of MSE and enables more balanced optimization across quantization stages. RMSE can be naturally defined over different feature groupings, yielding a unified family of four variants operating at the sample, channel, token, and element levels. These variants preserve the same cross-stage balancing property while applying gradient normalization within different feature groups at each stage.

Extensive experiments demonstrate the generality and practical benefits of replacing MSE with RMSE in learning-based PTQ: (1) RMSE consistently improves quantization performance in language modeling, commonsense reasoning, mathematical reasoning, and coding tasks across representative learning-based PTQ methods, model families, model scales, architectures, and quantization settings, with all four RMSE variants outperforming MSE; (2) the gains generally become more pronounced under aggressive quantization, with absolute average accuracy improvements ranging from $5.44\%$ to $10.83\%$ under W2A16 and from $0.75\%$ to $8.05\%$ under W4A4; and (3) as a drop-in replacement for MSE, RMSE requires no additional trainable parameters or hyperparameters while introducing negligible runtime and memory overhead. More importantly, our analysis and experimental results establish Optimization Imbalance as a previously overlooked phenomenon arising from the loss-dependent gradient scaling of MSE in sequential quantization and highlight optimization behavior as a fundamental design consideration for learning-based PTQ.

\section{Understanding Optimization Imbalance in Learning-based PTQ}

\subsection{Learning-based PTQ Preliminaries} 

We formulate learning-based PTQ under sequential quantization at the unit level, where a unit can be a layer, a block, or a group of consecutive blocks. Without loss of generality, we present the formulation for the more general weight-activation quantization setting, with weight-only quantization as the special case where activations remain in high precision. At the $i$-th stage, the $i$-th quantized unit is optimized to reconstruct the output of its high-precision counterpart. For the $i$-th unit, let $F_i$ denote its high-precision mapping with pre-trained weights $W_i$, and let $\widehat{F}_i$ denote its quantized counterpart. The quantized unit operates with effective quantized weights $\widehat{W}_i$ and quantized activations, governed by trainable auxiliary quantization parameters $\Theta_i$, such as scaling factors and rotation matrices. During reconstruction, $W_i$ remains frozen and only $\Theta_i$ is optimized. Given the high-precision input $X_i$ and quantized input $\widehat{X}_i$, the corresponding unit outputs are 
\begin{equation}
X_{i+1} = F_i(W_i, X_i), \qquad \widehat{X}_{i+1} = \widehat{F}_i(\widehat{W}_i, \widehat{X}_i; \Theta_i). 
\label{eq:fp_quantized_features} 
\end{equation} 
Most existing learning-based PTQ methods use mean squared error (MSE) as the reconstruction loss function. Specifically, the auxiliary parameters $\Theta_i$ are optimized by minimizing
\begin{equation} 
\mathcal{L}_{\mathrm{MSE}} = \frac{1}{N} \left\| \widehat{X}_{i+1} - X_{i+1} \right\|_F^2, 
\label{eq:mse} 
\end{equation} 
where $N$ denotes the total number of elements in $X_{i+1}$.

\subsection{Observation: Optimization Imbalance in Sequential Quantization}
\label{sec:optimization_imbalance}

While sequential quantization processes one unit at a time, it implicitly assumes that each unit receives sufficient optimization. But is this really the case? To answer this question, we systematically investigate optimization dynamics across units throughout the sequential quantization process, covering representative learning-based PTQ methods, model families, model scales, architectures, and quantization settings. A consistent pattern emerges across this broad spectrum: reconstruction loss magnitudes span a broad cross-stage range, which we term the \emph{reconstruction loss scale}, even though all units are optimized using the same MSE loss function. In many cases, the largest loss magnitude exceeds the smallest by more than $10{,}000\times$. Fig.~\ref{fig:method_main} presents a representative example, while comprehensive statistics are provided in Sec.~\ref{sec:main_results} and Appendix~\ref{appendix:optimization_statistics}. This broad loss scale arises naturally during sequential quantization. As quantization proceeds through the network, errors introduced by previously quantized units accumulate, making subsequent units increasingly difficult to reconstruct. Meanwhile, reconstruction losses are computed directly over unit outputs, whose magnitudes can vary substantially across units. Together, these factors give rise to the broad reconstruction loss scale observed across stages. 

This broad loss scale raises another question: \textit{do units with substantially different loss magnitudes receive comparable optimization strength?} We quantify this using the relative parameter update 
\begin{equation} 
\frac{\|\Delta \Theta_i\|_F}{\|\Theta_i\|_F}, 
\end{equation} 
computed over all trainable parameters at each optimization step and averaged across all steps in the quantization stage of the $i$-th unit. The relative parameter update measures how strongly a unit is modified during reconstruction and therefore serves as a practical proxy for the optimization strength received by that unit. 
We observe that the relative parameter update is consistently much larger for units with larger loss magnitudes, whereas units with smaller loss magnitudes receive only minimal updates. 
The close correspondence between loss magnitudes and relative parameter updates reveals a systematic imbalance in optimization across units, which we term \emph{Optimization Imbalance}. 
This imbalance is especially critical in sequential quantization, where each quantized unit remains fixed, so the errors left by an under-optimized unit propagate and accumulate through all subsequent units. Moreover, since loss magnitudes generally increase from early to late stages, early units tend to receive the weakest optimization, despite affecting the most subsequent units.

\subsection{Why MSE Leads to Optimization Imbalance} 
\label{sec:why_mse_lead_imbalance} 
What connects the reconstruction loss scale to Optimization Imbalance? Since gradients serve as the carriers of optimization signals, understanding this relationship requires analyzing how the reconstruction loss function influences the gradients received by each unit. A natural starting point is the parameter-space gradient, which directly determines parameter updates during reconstruction. Using the chain rule, we get
\begin{equation} 
\frac{\partial \mathcal L_{\mathrm{MSE}}} {\partial \Theta_{i}} = \frac{\partial \mathcal L_{\mathrm{MSE}}} {\partial \widehat X_{i+1}} \cdot \frac{\partial \widehat X_{i+1}} {\partial \Theta_{i}}. 
\end{equation} 
The parameter-space gradient is jointly determined by the loss function and the network Jacobian. Since our goal is to understand the optimization behavior induced by the loss function itself, independent of any particular network parameterization, we focus on $\partial \mathcal L_{\mathrm{MSE}} / \partial \widehat X_{i+1}$, which characterizes the output-space optimization signal generated directly by the loss function. This output-space perspective enables a cleaner theoretical analysis, while Sec.~\ref{sec:main_results} examines how the resulting gradient imbalance is reflected in parameter-space gradients and parameter updates. For the MSE loss defined in Eq.~\ref{eq:mse}, the output-space gradient magnitude is 
\begin{equation} 
\left\| \frac{\partial \mathcal L_{\mathrm{MSE}}} {\partial \widehat X_{i+1}} \right\|_F = \frac{2}{N} \left\| \widehat X_{i+1} - X_{i+1} \right\|_F = 2 \sqrt{ \frac{\mathcal L_{\mathrm{MSE}}}{N} }. 
\end{equation} 
This derivation reveals a fundamental property of MSE: the output-space gradient magnitude is proportional to the square root of the loss magnitude. Since the reconstruction loss scale can span more than $10{,}000\times$, this square-root scaling produces a range of output-space gradient magnitudes exceeding $100\times$. Thus, MSE inherently couples optimization strength to loss magnitude, directly accounting for Optimization Imbalance. Although the ultimate optimization dynamics are further influenced by network parameterization, this broad range of output-space gradient magnitudes provides a natural explanation for the highly uneven optimization strength observed in practice.

\section{Mitigating Optimization Imbalance via RMSE}

\subsection{RMSE for Balanced Optimization}
\label{sec:rmse_for_balanced_optimization}
Our analysis suggests a general principle for improving learning-based PTQ: an ideal loss function should decouple optimization strength across stages from the reconstruction loss scale. Root mean squared error (RMSE), a square-root transformation of MSE, naturally realizes this principle:
\begin{equation}
\mathcal{L}_{\mathrm{RMSE}}=\sqrt{\frac{1}{N}\left\|\widehat{X}_{i+1}-X_{i+1}\right\|_F^2}.
\end{equation}
Specifically, its output-space gradient magnitude is given by
\begin{equation} 
\left\| \frac{\partial \mathcal L_{\mathrm{RMSE}}}{\partial \widehat{X}_{i+1}} \right\|_F = \left\| \frac{1}{\sqrt{N}} \frac{\widehat{X}_{i+1}-X_{i+1}} {\left\|\widehat{X}_{i+1}-X_{i+1}\right\|_F} \right\|_F = \frac{1}{\sqrt{N}}, 
\end{equation} 
which is independent of the reconstruction loss magnitude. Compared with MSE, RMSE preserves the same gradient direction and relative magnitudes among gradient elements, while normalizing the overall gradient magnitude to a constant. 
In other words, although the RMSE loss itself may still span a broad range of magnitudes, this variation does not translate into imbalanced gradient magnitudes, thereby enabling balanced optimization despite the broad reconstruction loss scale. 
Importantly, this implicit gradient normalization emerges naturally from the loss function itself, requiring neither additional hyperparameters nor modifications to the learning-based PTQ pipeline.

\subsection{A Unified Family of RMSE Loss Functions} 
\label{sec:rmse_family} 
The analysis above establishes the gradient normalization property of RMSE, enabling balanced optimization across stages. The specific form of RMSE depends on which reconstruction errors are aggregated under each square root. Formally, let \(\{G_1,G_2,\ldots,G_M\}\) denote a partition of all \(N\) feature elements into \(M\) equal-sized groups. The generalized RMSE loss function is defined as 
\begin{equation} 
\mathcal{L}_{\mathrm{RMSE}}^{(G)} = \frac{1}{M} \sum_{m=1}^{M} \sqrt{ \frac{1}{|G_m|} \sum_{j\in G_m} \left( \widehat X_{i+1,j} - X_{i+1,j} \right)^2 }. 
\end{equation} 
For any such equal-sized grouping, the overall output-space gradient magnitude remains \(1/\sqrt{N}\), independent of both the reconstruction loss magnitude and the grouping strategy. Different groupings therefore preserve the same cross-stage balancing property while applying gradient normalization within different feature groups at each stage. For a reconstruction error tensor with sample, token, and channel dimensions, its elements can be naturally grouped at the sample, channel, token, or element level. These grouping strategies yield four RMSE variants. 

Specifically, \textbf{RMSE-Sample} treats each calibration sample as an individual group. \textbf{RMSE-Channel} treats each channel within each sample as an individual group. \textbf{RMSE-Token} treats each token within each sample as an individual group. \textbf{RMSE-Element} treats each feature element as an individual group and therefore reduces to the L1 loss. Their concrete formulations are provided in Appendix~\ref{appendix:rmse_variant_definitions}. Rather than independent heuristics, these variants represent different instantiations of the same gradient normalization principle. Consistent with their shared principle, all four variants outperform MSE across the evaluations in Sec.~\ref{sec:ablation_studies} and Appendix~\ref{appendix:rmse_variants}, while their relative performance differs across model families. We further observe that reconstruction loss magnitudes exhibit different degrees of within-stage variation under different feature groupings (Fig.~\ref{fig:groupwise_loss_distribution}), indicating that feature grouping constitutes a meaningful design dimension.

\section{Experiments}

\subsection{Experimental Setup}
\textbf{Models.} For evaluation, we select five representative LLMs spanning different model families, model scales, and architectures (dense and MoE): LLaMA2-7B and LLaMA3.1-8B from the LLaMA family~\citep{llama2,llama3}, and Qwen3.5-9B, Qwen3.5-27B, and Qwen3.5-35B-A3B from the Qwen3.5 family~\citep{qwen3.5}. Among them, LLaMA2-7B and Qwen3.5-9B serve as our primary evaluation models, representing a widely adopted PTQ benchmark and a modern hybrid linear-attention architecture, respectively.

\textbf{Quantization Methods.} We evaluate four representative learning-based PTQ methods spanning diverse learnable quantization designs: scaling factors and clipping thresholds in OmniQuant~\citep{omniquant}, orthogonal rotations and permutations in DuQuant~\citep{duquant}, affine transformations in FlatQuant~\citep{flatquant}, and scaling, low-rank adaptation, and sliding-window reconstruction in SliderQuant~\citep{sliderquant}. Despite their substantially different designs, all four methods rely on MSE-based feature reconstruction. This diversity allows us to examine whether Optimization Imbalance arises from the shared loss function rather than any particular quantization design, and whether RMSE generalizes across different learning-based PTQ methods. Following their official implementations, each unit corresponds to a transformer block, except for SliderQuant, which reconstructs a sliding window of consecutive blocks.

\textbf{Quantization Settings and Implementation Details.} We evaluate five quantization settings spanning weight-only and weight-activation quantization: W4A16, W3A16, W2A16, W2A16G128 (group size 128), and W4A4. Unless otherwise specified, we follow the official implementations and hyperparameter settings of each method and add support for model families not covered by the official code. For fair comparison, all results are obtained from our own reproductions, enabling a unified analysis of optimization behavior across different methods. We evaluate all methods using a unified LM-Evaluation-Harness~\citep{lm-eval-harness} pipeline, reporting WikiText2~\citep{wikitext2} perplexity (PPL) and zero-shot accuracy on PIQA~\citep{piqa}, ARC-Easy and ARC-Challenge~\citep{arc}, Winogrande~\citep{winogrande}, HellaSwag~\citep{hellaswag}, and BoolQ~\citep{boolq}. Unless otherwise specified, we use RMSE-Token for LLaMA models and RMSE-Channel for Qwen models based on their overall empirical performance. \textit{More implementation details are provided in Appendix~\ref{appendix:implementation}.}

\subsection{Main Results}
\label{sec:main_results}
\textbf{Generalization Across Models and Quantization Settings.} If Optimization Imbalance is indeed detrimental to learning-based PTQ, mitigating it should consistently improve quantization performance across diverse settings. We first test this hypothesis across model families, model scales, architectures, and quantization settings. We adopt OmniQuant as a representative framework because it covers both weight-only and weight-activation quantization, enabling controlled evaluation across diverse quantization regimes. As shown in Table~\ref{tab:across_models_and_settings}, replacing MSE with RMSE reduces perplexity and improves downstream accuracy in every evaluated configuration. While RMSE yields modest improvements under relatively mild settings such as W4A16, substantially larger gains emerge under more aggressive settings such as W2A16, W2A16G128, and W4A4. For example, on Qwen3.5-27B under W2A16 quantization, RMSE reduces perplexity from 37.51 to 22.20 while improving downstream accuracy from 60.45\% to 70.96\%. This trend suggests that Optimization Imbalance becomes increasingly detrimental as reconstruction becomes more difficult, making balanced optimization particularly important under aggressive quantization.

To relate these performance gains to the underlying optimization behavior, we examine the corresponding statistics in Fig.~\ref{fig:across_models_and_settings}. Under MSE, the broad loss scale is translated into highly uneven output-space gradients and further reflected in parameter-space gradients and relative parameter updates. In contrast, RMSE maintains constant output-space gradient magnitudes despite the broad loss scale, resulting in considerably more balanced gradients and updates across stages. Together, the performance gains and optimization statistics validate our analysis of Optimization Imbalance and establish balanced optimization as an effective design principle for learning-based PTQ.

\begin{table*}[t!] 
\vskip -0. in
\centering 
\caption{Generalization of RMSE under OmniQuant across diverse LLMs and quantization settings. Acc denotes the average accuracy over PIQA, ARC-Easy, ARC-Challenge, Winogrande, HellaSwag, and BoolQ.} 
\vskip -0.07 in
\renewcommand{\arraystretch}{1.1}
\setlength{\tabcolsep}{3.5pt}
\resizebox{\textwidth}{!}{ \begin{tabular}{l|c|cc|cc|cc|cc|cc|cc} \hline 
\multirow{2}{*}{Model} & \multirow{2}{*}{Loss}& \multicolumn{2}{c|}{Original} & \multicolumn{2}{c|}{W4A16} & \multicolumn{2}{c|}{W3A16} & \multicolumn{2}{c|}{W2A16} & \multicolumn{2}{c|}{W2A16G128} & \multicolumn{2}{c}{W4A4} \\
\cline{3-14} & & PPL$\downarrow$ & Acc$\uparrow$ & PPL$\downarrow$ & Acc$\uparrow$ & PPL$\downarrow$ & Acc$\uparrow$ & PPL$\downarrow$ & Acc$\uparrow$ & PPL$\downarrow$ & Acc$\uparrow$ & PPL$\downarrow$ & Acc$\uparrow$ \\
\hline 
\multirow{2}{*}{LLaMA2-7B} & MSE & \multirow{2}{*}{5.47} & \multirow{2}{*}{70.45} & 5.75 & 68.33 & 6.62 & 64.05 & 36.92 & 43.98 & 11.05 & 51.66 & 17.42 & 49.09 \\
& RMSE & & & 5.70 & 69.07 & 6.38 & 66.78 & 16.06 & 49.97 & 9.10 & 57.29 & 11.48 & 56.23 \\
\hline 
\multirow{2}{*}{LLaMA3.1-8B} & MSE & \multirow{2}{*}{6.24} & \multirow{2}{*}{75.04} & 7.19 & 72.59 & 12.38 & 60.55 & 1021.33 & 37.00 & 124.49 & 38.07 & 110.88 & 40.09 \\
& RMSE & & & 6.92 & 72.89 & 9.46 & 67.17 & 140.63 & 42.44 & 20.21 & 48.60 & 44.29 & 43.51 \\
\hline 
\multirow{2}{*}{Qwen3.5-9B} & MSE & \multirow{2}{*}{8.65} & \multirow{2}{*}{75.10} & 9.28 & 74.23 & 11.56 & 71.10 & 99.05 & 45.44 & 23.61 & 53.54 & 69.74 & 45.43 \\
& RMSE & & &  9.14 & 74.35 & 11.21 & 72.79 & 51.56 & 54.78 & 17.98 & 60.03 & 51.88 & 49.37 \\ \hline
\multirow{2}{*}{Qwen3.5-27B} & MSE & \multirow{2}{*}{6.90} & \multirow{2}{*}{77.10} & 7.17 & 77.71 & 8.42 & 77.12 & 37.51 & 60.45 & 11.73 & 70.88 & 29.45 & 59.46 \\
& RMSE & & & 7.09 & 78.37 & 8.26 & 78.13 & 22.20 & 70.96 & 11.29 & 73.09 & 16.13 & 67.51 \\ 
\hline 
\multirow{2}{*}{Qwen3.5-35B-A3B} & MSE & \multirow{2}{*}{6.59} & \multirow{2}{*}{78.34} & 7.22 & 77.61 & 9.87 & 73.05 & 73.79 & 46.76 & 24.66 & 56.47 & 12.42 & 65.63 \\
& RMSE & & & 7.06 & 77.98 & 8.83 & 74.50 & 21.85 & 57.59 & 13.90 & 66.76 & 10.82 & 67.48 \\ \hline
\end{tabular} } 
\label{tab:across_models_and_settings} 
\vskip 0.25 in

\centering
\centerline{\includegraphics[width=0.99\textwidth]{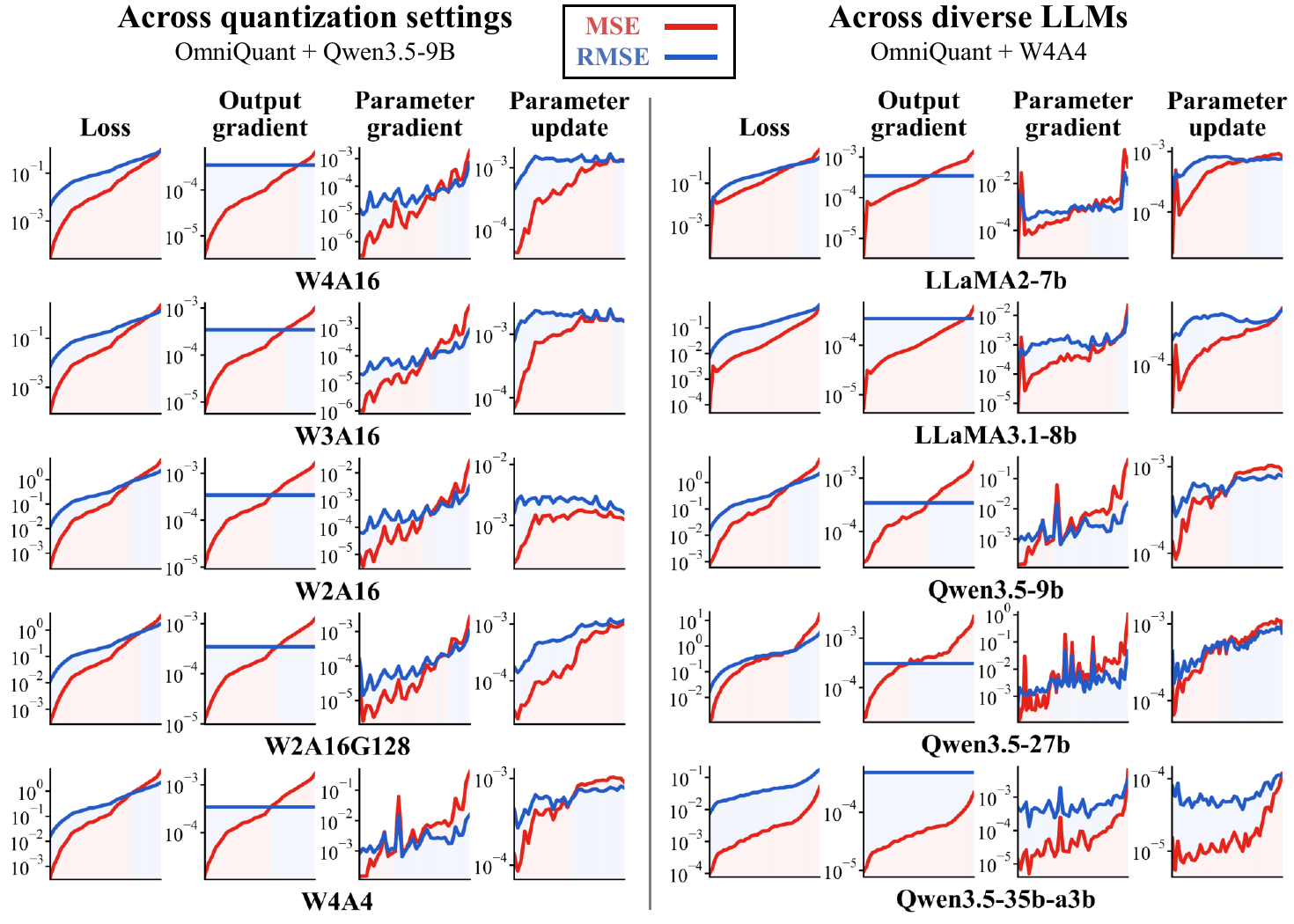}}
\captionof{figure}{Optimization statistics under MSE and RMSE across quantization settings and diverse LLMs. The horizontal axis represents the quantization stages from early to late. For each stage, loss, output-space gradient, parameter-space gradient, and relative parameter update are averaged over all optimization steps. Parameter-space gradients and parameter updates are computed over trainable parameters only.}
\label{fig:across_models_and_settings}
\vskip -0.0 in
\end{table*}

\clearpage

\begin{figure*}[t] 
\centering 
\vskip -0.2 in
\begin{minipage}[t]{0.53\textwidth} 
\captionsetup{type=table} 
\caption{Generalization of RMSE across representative learning-based PTQ methods under W4A4 quantization on LLaMA2-7B and Qwen3.5-9B. Acc denotes the average accuracy over PIQA, ARC-Easy, ARC-Challenge, Winogrande, HellaSwag, and BoolQ.} 
\vskip -0.02 in
\label{tab:across_methods} 
\centering 
\renewcommand{\arraystretch}{1.1} \resizebox{\linewidth}{!}{ 
\begin{tabular}{c|l|c|cc} 
\hline Model & Method & Loss & PPL$\downarrow$ & Acc$\uparrow$ \\
\hline \multirow{9}{*}{LLaMA2-7B} & Original & -- & 5.47 & 70.45 \\
\cline{2-5} & \multirow{2}{*}{OmniQuant} & MSE & 17.42 & 49.09 \\
& & RMSE & 11.48 & 56.23 \\
\cline{2-5} & \multirow{2}{*}{SliderQuant} & MSE & 8.48 & 57.20 \\
& & RMSE & 8.14 & 59.74 \\
\cline{2-5} & \multirow{2}{*}{DuQuant} & MSE & 6.07 & 66.91 \\
& & RMSE & 6.01 & 67.69 \\
\cline{2-5} & \multirow{2}{*}{FlatQuant} & MSE & 5.82 & 67.72 \\
& & RMSE & 5.78 & 68.76 \\
\hline \multirow{9}{*}{Qwen3.5-9B} & Original & -- & 8.65 & 75.10 \\
\cline{2-5} & \multirow{2}{*}{OmniQuant} & MSE & 69.74 & 45.43 \\
& & RMSE & 51.88 & 49.37 \\
\cline{2-5} & \multirow{2}{*}{SliderQuant} & MSE & 21.41 & 60.43 \\
& & RMSE & 20.64 & 61.92 \\
\cline{2-5} & \multirow{2}{*}{DuQuant} & MSE & 10.95 & 69.85 \\
& & RMSE & 10.54 & 70.60 \\
\cline{2-5} & \multirow{2}{*}{FlatQuant} & MSE & 10.26 & 72.23 \\
& & RMSE & 9.15 & 74.19 \\
\hline 
\end{tabular} } 
\end{minipage} 
\hfill 
\begin{minipage}[t]{0.45\textwidth} 
\captionsetup{type=figure} 
\captionsetup{skip=2pt}
\centering 
\includegraphics[width=\linewidth]{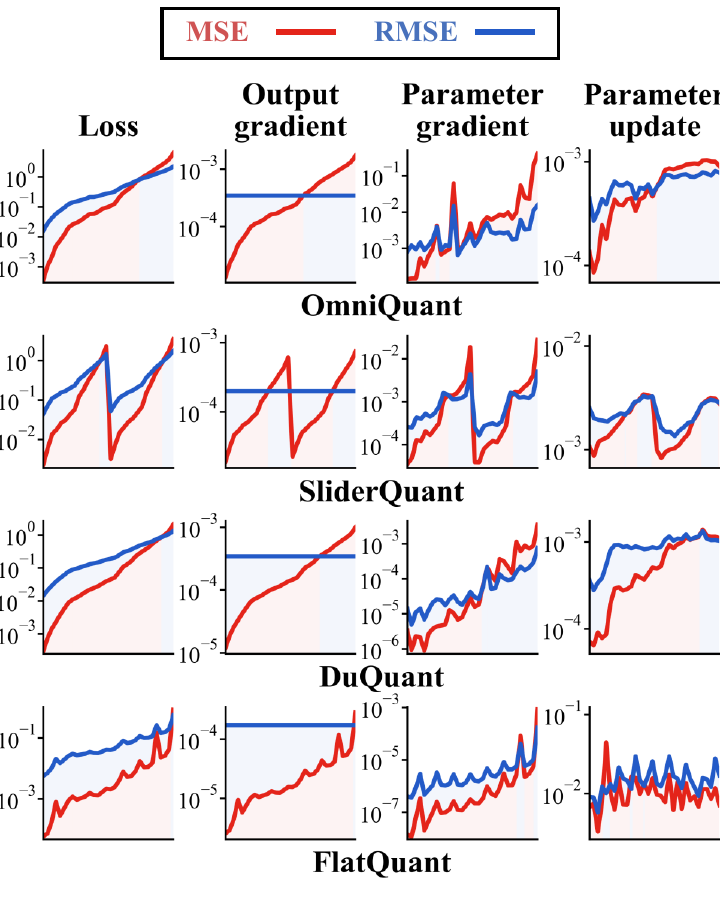} 
\caption{Optimization statistics under MSE and RMSE across representative learning-based PTQ methods on Qwen3.5-9B. The horizontal axis denotes the quantization stages from early to late.} 
\label{fig:across_methods} 
\end{minipage} 
\end{figure*}

\textbf{Generalization Across PTQ Methods.} We further test whether the preceding findings generalize beyond OmniQuant. Specifically, we extend our analysis to three additional representative learning-based PTQ methods, namely SliderQuant, DuQuant, and FlatQuant, which employ substantially different quantization designs. As shown in Table~\ref{tab:across_methods}, RMSE yields consistent performance gains across all four methods despite their distinct designs. The corresponding statistics in Fig.~\ref{fig:across_methods} reveal the same optimization pattern: under MSE, the broad reconstruction loss scale is accompanied by uneven gradients and parameter updates, whereas RMSE produces considerably more balanced optimization. These results demonstrate that Optimization Imbalance is broadly shared across representative learning-based PTQ methods. Its consistent presence across different quantization designs further supports the loss-dependent gradient scaling of MSE as its source.

\subsection{Ablation Studies}
\label{sec:ablation_studies}
\begin{table*}[t] 
\centering 
\caption{Comparison of RMSE variants across quantization settings under OmniQuant. Results are reported as average accuracy (\%) on downstream benchmarks.} 
\label{tab:rmse_variants} 
\renewcommand{\arraystretch}{1.2} \setlength{\tabcolsep}{4pt} 
\resizebox{0.95\textwidth}{!}{ \begin{tabular}{c|c|cccc|c|cccc} \hline 
\multirow{3}{*}{\#Bits} & \multicolumn{5}{c|}{LLaMA2-7B} & \multicolumn{5}{c}{Qwen3.5-9B} \\ \cline{2-11}
& \multirow{2}{*}{MSE} & \multicolumn{4}{c|}{RMSE} & \multirow{2}{*}{MSE} & \multicolumn{4}{c}{RMSE} \\
&  & Sample & Channel & Token & Element &  & Sample & Channel & Token & Element \\
\hline W4A16 & 68.33 & 68.88 & \textbf{69.33} & 69.07 & 68.64 & 74.23 & 74.97 & 74.35 & \textbf{75.08} & 74.70 \\
W3A16 & 64.05 & 64.17 & 65.88 & \textbf{66.78} & 66.58 & 71.10 & \textbf{73.17} & 72.79 & 72.47 & 72.92 \\
W2A16 & 43.98 & 45.09 & 46.01 & \textbf{49.97} & 46.76 & 45.44 & 47.61 & \textbf{54.78} & 47.63 & 52.76 \\
W2A16G128 & 51.66 & 56.58 & 57.36 & 57.29 & \textbf{57.54} & 53.54 & \textbf{61.08} & 60.03 & 59.89 & 58.63 \\
W4A4 & 49.09 & 51.73 & 53.40 & \textbf{56.23} & 54.35 & 45.43 & 47.85 & \textbf{49.37} & 47.38 & 48.79 \\
\hline 
\end{tabular}} 
\end{table*}

\textbf{Comparing RMSE Variants.} As introduced in Sec.~\ref{sec:rmse_family}, all four RMSE variants preserve the same cross-stage balancing property while applying gradient normalization to different feature groups within each unit. As shown in Table~\ref{tab:rmse_variants}, all four variants consistently outperform MSE across different models and quantization settings, confirming the effectiveness of this principle across feature groupings. At the same time, their relative performance varies across model families: LLaMA-family models generally favor RMSE-Token, whereas Qwen-family models generally favor RMSE-Channel. This model-dependent preference further shows that feature grouping constitutes a meaningful design dimension for balanced optimization. We therefore adopt RMSE-Token for LLaMA models and RMSE-Channel for Qwen models throughout the main experiments. For brevity, we present a subset of results here, \textit{while comprehensive comparisons across additional models, quantization settings, and learning-based PTQ methods are provided in Appendix~\ref{appendix:rmse_variants}}.

\textbf{Generalization to Reasoning Tasks.} To examine whether balanced optimization benefits complex reasoning, we evaluate the quantized models obtained under OmniQuant, without task-specific training or adaptation, on mathematical reasoning (GSM8K~\citep{gsm8k} and MATH-500~\citep{math-500}) and coding (HumanEval~\citep{humaneval} and MBPP~\citep{mbpp}). As shown in Table~\ref{tab:reasoning_tasks}, RMSE consistently improves overall performance across quantization settings. Under W2A16G128, however, RMSE alone cannot fully recover the severely degraded performance, highlighting the need for specialized quantization designs in such aggressive settings.

\begin{table*}[t]
\centering 
\caption{Generalization of RMSE to mathematical reasoning and coding tasks under OmniQuant on Qwen3.5-9B. All results are reported as accuracy (\%).}
\vskip -0.1 in
\label{tab:reasoning_tasks} 
\renewcommand{\arraystretch}{1.1} 
\setlength{\tabcolsep}{4pt} 
\resizebox{0.75\linewidth}{!}{ \begin{tabular}{c|c|cc|cc|c} \hline \multirow{2}{*}{\#Bits} & \multirow{2}{*}{Loss} & \multicolumn{2}{c|}{Mathematical Reasoning} & \multicolumn{2}{c|}{Coding} & \multirow{2}{*}{Avg} \\
\cline{3-6} & & GSM8K & MATH-500 & HumanEval & MBPP & \\ \hline
Original & -- & 96.44 & 97.20 & 85.37 & 70.37 & 87.35 \\
\hline \multirow{2}{*}{W4A16} & MSE & 95.15 & 94.20 & 75.61 & 57.14 & 80.53 \\
& RMSE & 95.98 & 94.80 & 75.00 & 62.96 & \textbf{82.19} \\
\hline \multirow{2}{*}{W3A16} & MSE & 78.70 & 51.20 & 32.93 & 24.34 & 46.79 \\
& RMSE & 81.65 & 54.60 & 35.98 & 25.40 & \textbf{49.41} \\
\hline \multirow{2}{*}{W2A16G128} & MSE & 1.67 & 2.20 & 0.00 & 0.00 & 0.97 \\
& RMSE & 7.35 & 7.00 & 0.00 & 0.26 & \textbf{3.65} \\
\hline 
\end{tabular}} 
\vskip 0.05 in
\end{table*}

\begin{figure*}[t!] 
\centering 
\centerline{\includegraphics[width=0.99\textwidth]{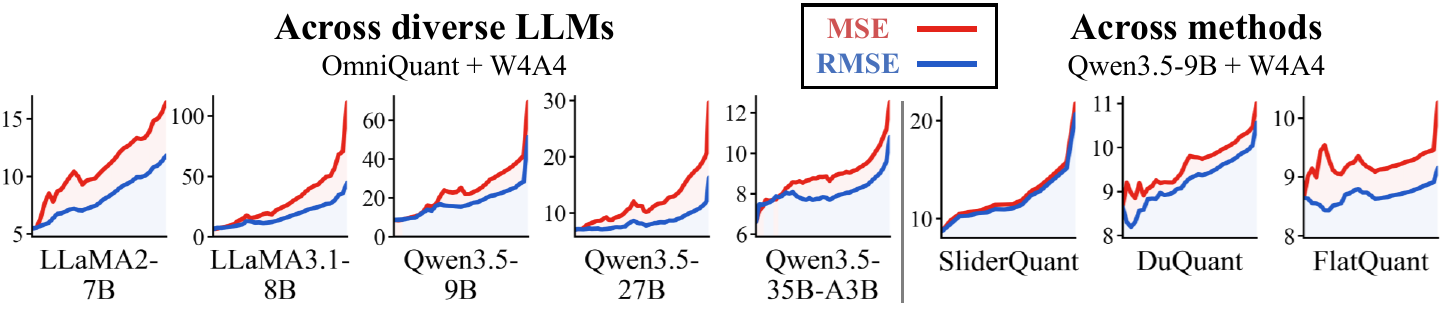}} 
\vskip -0.1 in 
\caption{Perplexity evolution with quantization depth under MSE and RMSE. The horizontal axis represents the quantization stages from early to late. Left: OmniQuant with W4A4 quantization across diverse LLMs. Right: representative learning-based PTQ methods on Qwen3.5-9B under W4A4 quantization.}
\label{fig:ppl_depth} 
\vskip 0.05 in
\end{figure*}

\begin{table}[t!] 
\centering 
\caption{Quantization process efficiency comparison between MSE and RMSE under OmniQuant. Runtime and peak GPU memory are measured over the entire sequential quantization process on a single NVIDIA L40 GPU with a batch size of 1.}
\label{tab:efficiency} 
\vskip -0.08 in 
\renewcommand{\arraystretch}{1.1} \setlength{\tabcolsep}{4pt} \resizebox{0.7\linewidth}{!}{ \begin{tabular}{l|c|ccc|ccc} 
\hline 
\multirow{2}{*}{Model} & \multirow{2}{*}{\#Bits} & \multicolumn{3}{c|}{Runtime (minutes)} & \multicolumn{3}{c}{Peak memory (GB)} \\
\cline{3-8} & & MSE & RMSE & Overhead & MSE & RMSE & Overhead \\
\hline \multirow{2}{*}{LLaMA2-7B} & W4A16 & 123.9 & 127.4 & +2.82\% & 6.14 & 6.14 & +0.00\% \\
& W4A4 & 181.7 & 181.1 & -0.33\% & 7.28 & 7.28 & +0.00\% \\
\hline \multirow{2}{*}{Qwen3.5-9B} & W4A16 & 184.8 & 182.6 & -1.19\% & 5.63 & 5.63 & +0.00\% \\
& W4A4 & 233.9 & 236.3 & +1.02\% & 6.90 & 6.90 & +0.00\% \\
\hline 
\end{tabular}} 
\vskip -0. in
\end{table}

\textbf{Perplexity Across Quantization Depth.} To further connect optimization behavior with model performance, we record the perplexity of partially quantized models throughout sequential quantization. As shown in Fig.~\ref{fig:ppl_depth}, RMSE consistently maintains lower perplexity than MSE, with the gap generally widening as quantization progresses. This trend highlights the increasing importance of balanced optimization as quantization errors accumulate.

\textbf{Computational Overhead.} As shown in Table~\ref{tab:efficiency}, RMSE introduces negligible runtime and memory overhead as a drop-in replacement for MSE, enabling more balanced optimization while preserving computational efficiency.

\section{Discussion} 
Existing learning-based PTQ research has primarily focused on developing advanced quantization designs. Our work suggests a complementary perspective: the choice of loss function itself fundamentally shapes optimization behavior across stages during sequential quantization. The consistent benefits of RMSE, particularly under aggressive quantization, indicate that Optimization Imbalance is not a minor artifact but a systematic phenomenon arising from the loss-dependent gradient scaling of MSE during sequential quantization. Its impact becomes increasingly pronounced as reconstruction becomes more difficult. This issue is particularly consequential in sequential quantization, where insufficient optimization at an early stage may influence all subsequent quantization stages. Furthermore, the distinct behavior of different feature grouping strategies indicates that feature grouping constitutes a meaningful design dimension, opening a broader space for balanced optimization beyond the variants studied here. Taken together, these findings suggest that optimization behavior should be treated as a fundamental design consideration in future learning-based PTQ research, alongside quantization design.

\section{Conclusion} 
We identify \textit{Optimization Imbalance} as a previously overlooked phenomenon arising from MSE-based reconstruction during sequential quantization in learning-based PTQ, where optimization strength across quantization stages is coupled with the reconstruction loss scale. To address this, we propose a family of RMSE loss functions that naturally decouple these two quantities and enable balanced optimization across quantization stages. These loss functions serve as drop-in replacements for MSE. Extensive experiments across learning-based PTQ methods, model families, model scales, architectures, and quantization settings demonstrate consistent improvements in quantization performance across language modeling, commonsense reasoning, mathematical reasoning, and coding tasks, with the gains becoming increasingly pronounced under more aggressive quantization.

\section*{Reproducibility Statement} 
To facilitate reproducibility, all experiments are conducted using publicly available models, datasets, and open-source PTQ frameworks, including OmniQuant, FlatQuant, DuQuant, and SliderQuant. Implementation details regarding our four RMSE variants based on different feature grouping strategies, experimental settings, hyperparameters, and evaluation protocols are provided in the main paper and appendix. We will release our implementation and experimental code upon publication.

\section*{AI Use Statement} 
We used generative AI tools solely to assist with proofreading the manuscript. All AI-assisted content was reviewed and verified by the authors. The authors take full responsibility for the content of this work.

\bibliographystyle{iclr2027_conference}
\bibliography{iclr2027_conference}

@inproceedings{omniquant,
  title={OmniQuant: Omnidirectionally Calibrated Quantization for Large Language Models},
  author={Shao, Wenqi and Chen, Mengzhao and Zhang, Zhaoyang and Xu, Peng and Zhao, Lirui and Li, Zhiqian and Zhang, Kaipeng and Gao, Peng and Qiao, Yu and Luo, Ping},
  booktitle={ICLR},
  year={2024}
}

@inproceedings{quarot,
  title={QuaRot: Outlier-Free 4-Bit Inference in Rotated LLMs},
  author={Ashkboos, Saleh and Mohtashami, Amirkeivan and Croci, Maximilian L and Li, Bo and Cameron, Pashmina and Jaggi, Martin and Alistarh, Dan and Hoefler, Torsten and Hensman, James},
  booktitle={NeurIPS},
  year={2024}
}

@inproceedings{wikitext2,
  title={Pointer sentinel mixture models},
  author={Merity, Stephen and Xiong, Caiming and Bradbury, James and Socher, Richard},
  booktitle={ICLR},
  year={2017}
}

@inproceedings{transformer,
  title={Attention Is All You Need},
  author={Vaswani, Ashish and Shazeer, Noam and Parmar, Niki and Uszkoreit, Jakob and Jones, Llion and Gomez, Aidan N. and Kaiser, {\L}ukasz and Polosukhin, Illia},
  booktitle={NIPS},
  year={2017}
}

@article{deepseekv3,
  title={Deepseek-v3 technical report},
  author={Liu, Aixin and Feng, Bei and Xue, Bing and Wang, Bingxuan and Wu, Bochao and Lu, Chengda and Zhao, Chenggang and Deng, Chengqi and Zhang, Chenyu and Ruan, Chong and others},
  journal={arXiv preprint arXiv:2412.19437},
  year={2024}
}

@article{deepseek-r1,
  title={DeepSeek-R1 incentivizes reasoning in LLMs through reinforcement learning},
  author={Guo, Daya and Yang, Dejian and Zhang, Haowei and Song, Junxiao and Wang, Peiyi and Zhu, Qihao and Xu, Runxin and Zhang, Ruoyu and Ma, Shirong and Bi, Xiao and others},
  journal={Nature},
  volume={645},
  number={8081},
  pages={633--638},
  year={2025},
  publisher={Nature Publishing Group UK London}
}

@article{gemini,
  title={Gemini: A family of highly capable multimodal models},
  author={Anil, Rohan and Borgeaud, Sebastian and Wu, Yonghui and Alayrac, Jean-Baptiste and Yu, Jiahui and Soricut, Radu and Schalkwyk, Johan and Dai, Andrew M and Hauth, Anja and Millican, Katie and others},
  journal={arXiv preprint arXiv:2312.11805},
  year={2023}
}

@inproceedings{gpt3,
  title={Language models are few-shot learners},
  author={Brown, Tom and Mann, Benjamin and Ryder, Nick and Subbiah, Melanie and Kaplan, Jared D and Dhariwal, Prafulla and Neelakantan, Arvind and Shyam, Pranav and Sastry, Girish and Askell, Amanda and others},
  booktitle={NeurIPS},  
  year={2020}
}

@article{gpt4,
  title={Gpt-4 technical report},
  author={Achiam, Josh and Adler, Steven and Agarwal, Sandhini and Ahmad, Lama and Akkaya, Ilge and Aleman, Florencia Leoni and Almeida, Diogo and Altenschmidt, Janko and Altman, Sam and Anadkat, Shyamal and others},
  journal={arXiv preprint arXiv:2303.08774},
  year={2023}
}

@article{llama2,
  title={Llama 2: Open foundation and fine-tuned chat models},
  author={Touvron, Hugo and Martin, Louis and Stone, Kevin and Albert, Peter and Almahairi, Amjad and Babaei, Yasmine and Bashlykov, Nikolay and Batra, Soumya and Bhargava, Prajjwal and Bhosale, Shruti and others},
  journal={arXiv preprint arXiv:2307.09288},
  year={2023}
}

@article{llama3,
  title={The llama 3 herd of models},
  author={Grattafiori, Aaron and Dubey, Abhimanyu and Jauhri, Abhinav and Pandey, Abhinav and Kadian, Abhishek and Al-Dahle, Ahmad and Letman, Aiesha and Mathur, Akhil and Schelten, Alan and Vaughan, Alex and others},
  journal={arXiv preprint arXiv:2407.21783},
  year={2024}
}

@inproceedings{awq,
  title={AWQ: Activation-aware Weight Quantization for LLM Compression and Acceleration},
  author={Lin, Ji and Tang, Jiaming and Tang, Haotian and Yang, Shang and Chen, Wei-Ming and Wang, Wei-Chen and Xiao, Guangxuan and Dang, Xingyu and Gan, Chuang and Han, Song},
  booktitle={MLSys},
  year={2024}
}

@inproceedings{gptq,
  title={Gptq: Accurate post-training quantization for generative pre-trained transformers},
  author={Frantar, Elias and Ashkboos, Saleh and Hoefler, Torsten and Alistarh, Dan},
  booktitle={ICLR},
  year={2023}
}

@inproceedings{cbq,
  title={Cbq: Cross-block quantization for large language models},
  author={Ding, Xin and Liu, Xiaoyu and Tu, Zhijun and Zhang, Yun and Li, Wei and Hu, Jie and Chen, Hanting and Tang, Yehui and Xiong, Zhiwei and Yin, Baoqun and others},
  booktitle={ICLR},
  year={2025}
}

@inproceedings{duquant,
  title={DuQuant: Distributing Outliers via Dual Transformation Makes Stronger Quantized LLMs},
  author={Lin, Haokun and Xu, Haobo and Wu, Yichen and Cui, Jingzhi and Zhang, Yingtao and Mou, Linzhan and Song, Linqi and Sun, Zhenan and Wei, Ying},
  booktitle={NeurIPS},
  year={2024}
}

@inproceedings{smoothquant,
  title={Smoothquant: Accurate and efficient post-training quantization for large language models},
  author={Xiao, Guangxuan and Lin, Ji and Seznec, Mickael and Wu, Hao and Demouth, Julien and Han, Song},
  booktitle={ICML},
  year={2023},
}

@inproceedings{spinquant,
  title={SpinQuant: LLM quantization with learned rotations},
  author={Liu, Zechun and Zhao, Changsheng and Fedorov, Igor and Soran, Bilge and Choudhary, Dhruv and Krishnamoorthi, Raghuraman and Chandra, Vikas and Tian, Yuandong and Blankevoort, Tijmen},
  booktitle={ICLR},
  year={2025},
}

@inproceedings{flatquant,
  title={Flatquant: Flatness matters for llm quantization},
  author={Sun, Yuxuan and Liu, Ruikang and Bai, Haoli and Bao, Han and Zhao, Kang and Li, Yuening and Hu, Jiaxin and Yu, Xianzhi and Hou, Lu and Yuan, Chun and others},
  booktitle={ICML},
  year={2025}
}

@article{bitnet-1.58,
  title={The era of 1-bit llms: All large language models are in 1.58 bits},
  author={Ma, Shuming and Wang, Hongyu and Ma, Lingxiao and Wang, Lei and Wang, Wenhui and Huang, Shaohan and Dong, Li and Wang, Ruiping and Xue, Jilong and Wei, Furu},
  journal={arXiv preprint arXiv:2402.17764},
  year={2024}
}

@inproceedings{onebit,
  title={OneBit: Towards Extremely Low-bit Large Language Models},
  author={Xu, Yuzhuang and Han, Xu and Yang, Zonghan and Wang, Shuo and Zhu, Qingfu and Liu, Zhiyuan and Liu, Weidong and Che, Wanxiang},
  booktitle={NeurIPS},
  year={2024}
}

@inproceedings{q8bert,
  title={Q8bert: Quantized 8bit bert},
  author={Zafrir, Ofir and Boudoukh, Guy and Izsak, Peter and Wasserblat, Moshe},
  booktitle={EMC2-NeurIPS},
  year={2019},
}

@inproceedings{piqa,
  title={Piqa: Reasoning about physical commonsense in natural language},
  author={Yonatan Bisk and Rowan Zellers and Ronan Le Bras and Jianfeng Gao and Yejin Choi},
  booktitle={AAAI},
  year={2020}
}

@inproceedings{hellaswag,
    title={HellaSwag: Can a Machine Really Finish Your Sentence?},
    author={Zellers, Rowan and Holtzman, Ari and Bisk, Yonatan and Farhadi, Ali and Choi, Yejin},
    booktitle ={ACL},
    year={2019}
}

@article{arc,
  title={Think you have solved question answering? try arc, the ai2 reasoning challenge},
  author={Clark, Peter and Cowhey, Isaac and Etzioni, Oren and Khot, Tushar and Sabharwal, Ashish and Schoenick, Carissa and Tafjord, Oyvind},
  journal={arXiv preprint arXiv:1803.05457},
  year={2018}
}

@article{winogrande,
  title={Winogrande: An adversarial winograd schema challenge at scale},
  author={Sakaguchi, Keisuke and Bras, Ronan Le and Bhagavatula, Chandra and Choi, Yejin},
  journal={Communications of the ACM},
  year={2021},
}

@inproceedings{boolq,
  title={BoolQ: Exploring the Surprising Difficulty of Natural Yes/No Questions},
  author={Clark, Christopher and Lee, Kenton and Chang, Ming-Wei and Kwiatkowski, Tom and Collins, Michael and Toutanova, Kristina},
  booktitle={NAACL},
  year={2019},
}

@inproceedings{math-500,
  title={Measuring mathematical problem solving with the math dataset},
  author={Hendrycks, Dan and Burns, Collin and Kadavath, Saurav and Arora, Akul and Basart, Steven and Tang, Eric and Song, Dawn and Steinhardt, Jacob},
  booktitle={NeurIPS},
  year={2021}
}

@inproceedings{evalplus,
  title={Is your code generated by chatgpt really correct? rigorous evaluation of large language models for code generation},
  author={Liu, Jiawei and Xia, Chunqiu Steven and Wang, Yuyao and Zhang, Lingming},
  booktitle={NeurIPS},
  year={2023}
}

@article{gsm8k,
  title={Training verifiers to solve math word problems},
  author={Cobbe, Karl and Kosaraju, Vineet and Bavarian, Mohammad and Chen, Mark and Jun, Heewoo and Kaiser, Lukasz and Plappert, Matthias and Tworek, Jerry and Hilton, Jacob and Nakano, Reiichiro and others},
  journal={arXiv preprint arXiv:2110.14168},
  year={2021}
}

@misc{qwen3.5,
    title = {{Qwen3.5}: Towards Native Multimodal Agents},
    url = {https://qwen.ai/blog?id=qwen3.5},
    author = {{Qwen Team}},
    year = {2026}
}

@inproceedings{sliderquant,
  title={SliderQuant: Accurate Post-Training Quantization for LLMs},
  author={Wang, Shigeng and  Li, Chao and  Kang, Yangyuxuan and  Fan, Jiawei and  Ou, Zhonghong and Yao, Anbang},
  booktitle={ICLR},
  year={2026},
}

@misc{lm-eval-harness,
  author       = {Gao, Leo and Tow, Jonathan and Abbasi, Baber and Biderman, Stella and Black, Sid and DiPofi, Anthony and Foster, Charles and Golding, Laurence and Hsu, Jeffrey and Le Noac'h, Alain and Li, Haonan and McDonell, Kyle and Muennighoff, Niklas and Ociepa, Chris and Phang, Jason and Reynolds, Laria and Schoelkopf, Hailey and Skowron, Aviya and Sutawika, Lintang and Tang, Eric and Thite, Anish and Wang, Ben and Wang, Kevin and Zou, Andy},
  title        = {The Language Model Evaluation Harness},
  year         = {2024},
  url          = {https://zenodo.org/records/12608602}
}

@inproceedings{gptaq,
  title={{GPTAQ:} Efficient Finetuning-Free Quantization for Asymmetric Calibration},
  author={Li, Yuhang and Yin, Ruokai and Lee, Donghyun and Xiao, Shiting and Panda, Priyadarshini},
  booktitle={ICML},
  year={2025},
}

@inproceedings{perq,
  title={Pushing the limits of block rotations in post-training quantization},
  author={Sanjeet, Sai and Colbert, Ian and Monteagudo-Lago, Pablo and Franco, Giuseppe and Umuroglu, Yaman and Fraser, Nicholas J},
  booktitle={ICML},
  year={2026}
}

@inproceedings{wush,
  title={Wush: Near-optimal adaptive transforms for llm quantization},
  author={Chen, Jiale and Egiazarian, Vage and Castro, Roberto L and Hoefler, Torsten and Alistarh, Dan},
  booktitle={ICML},
  year={2026}
}

@inproceedings{osaq,
  title={OSAQ: Outlier Self-Absorption for Accurate Low-bit LLM Quantization},
  author={Li, Zhikai and Dong, Zhen and Liu, Xuewen and Zhang, Jing and Gu, Qingyi},
  booktitle={ICML},
  year={2026}
}

@inproceedings{fptquant,
  title={Fptquant: Function-preserving transforms for llm quantization},
  author={van Breugel, Boris and Bondarenko, Yelysei and Whatmough, Paul and Nagel, Markus},
  booktitle={ICML},
  year={2026}
}

@inproceedings{paroquant,
  title={Paroquant: Pairwise rotation quantization for efficient reasoning llm inference},
  author={Liang, Yesheng and Chen, Haisheng and Han, Song and Liu, Zhijian},
  booktitle={ICLR},
  year={2026}
}

@inproceedings{lrquant,
  title={LRQuant: Learnable and robust post-training quantization for large language models},
  author={Zhao, Jiaqi and Zhang, Miao and Zeng, Chao and Wang, Ming and Liu, Xuebo and Nie, Liqiang},
  booktitle={ACL},
  year={2024}
}

@article{humaneval,
  title={Evaluating large language models trained on code},
  author={Chen, Mark and Tworek, Jerry and Jun, Heewoo and Yuan, Qiming and Pinto, Henrique Ponde De Oliveira and Kaplan, Jared and Edwards, Harri and Burda, Yuri and Joseph, Nicholas and Brockman, Greg and others},
  journal={arXiv preprint arXiv:2107.03374},
  year={2021}
}

@article{mbpp,
  title={Program synthesis with large language models},
  author={Austin, Jacob and Odena, Augustus and Nye, Maxwell and Bosma, Maarten and Michalewski, Henryk and Dohan, David and Jiang, Ellen and Cai, Carrie and Terry, Michael and Le, Quoc and others},
  journal={arXiv preprint arXiv:2108.07732},
  year={2021}
}

\newpage

\section*{Appendix}
\appendix

\section{Implementation Details}
\label{appendix:implementation}

\subsection{Experimental Details} 
\label{appendix:experimental_details}

We follow the official implementations and hyperparameter settings of each PTQ method. For a fair comparison, RMSE is used as a drop-in replacement for the original MSE loss function, while all other components, including trainable parameters, optimizer, learning rate, reconstruction steps, and quantization configurations, are kept unchanged. When an official implementation does not directly support a model family, we implement the corresponding quantization pipeline following the original method design. All methods are calibrated using 128 samples from WikiText2, each with a sequence length of 2048, following the standard protocol used in prior PTQ work. For each MSE/RMSE comparison, we use exactly the same calibration samples to ensure that the observed differences come from the loss function rather than data variation.

For reasoning-intensive evaluation, we evaluate GSM8K and MATH-500 using LM-Evaluation-Harness, and HumanEval and MBPP using EvalPlus~\citep{evalplus}. We adopt the officially recommended sampling settings of Qwen3.5 for reasoning generation. All reasoning evaluations are conducted on the same quantized models without task-specific training or adaptation.

All quantization experiments are conducted on a single NVIDIA L40 GPU. For larger models, including Qwen3.5-27B and Qwen3.5-35B-A3B, evaluation is performed using two L40 GPUs to accommodate model inference. The runtime and peak memory reported in Table~\ref{tab:efficiency} are measured during reconstruction on a single NVIDIA L40 GPU with a batch size of 1.

\subsection{Definitions of RMSE Variants} 
\label{appendix:rmse_variant_definitions} 
We provide the concrete definitions of the RMSE variants used in our experiments. Let $E = \widehat{X}_{i+1} - X_{i+1} \in \mathbb{R}^{B \times T \times C}$ denote the reconstruction error tensor at a quantization stage, where $B$, $T$, and $C$ denote the batch size, number of tokens, and number of channels, respectively. The official implementations adopt different reconstruction batch sizes (\(B=1\) for OmniQuant and DuQuant, \(B=4\) for FlatQuant, and \(B=3\) for SliderQuant). We therefore define all grouping strategies within each sample, making the sample level the coarsest formulation shared consistently across methods. 
\paragraph{RMSE-Sample.} RMSE-Sample computes RMSE independently for each calibration sample and then averages over samples: 
\begin{equation} 
\mathcal{L}_{\mathrm{Sample}} = \frac{1}{B} \sum_{b=1}^{B} \sqrt{ \frac{1}{TC} \sum_{t=1}^{T} \sum_{c=1}^{C} E_{b,t,c}^{2} }. 
\end{equation} 
\paragraph{RMSE-Channel.} RMSE-Channel computes RMSE independently for each channel within each sample and then averages over all sample-channel pairs: 
\begin{equation} 
\mathcal{L}_{\mathrm{Channel}} = \frac{1}{BC} \sum_{b=1}^{B} \sum_{c=1}^{C} \sqrt{ \frac{1}{T} \sum_{t=1}^{T} E_{b,t,c}^{2} }. 
\end{equation} 
\paragraph{RMSE-Token.} RMSE-Token computes RMSE independently for each token and then averages over all tokens: \begin{equation} 
\mathcal{L}_{\mathrm{Token}} = \frac{1}{BT} \sum_{b=1}^{B} \sum_{t=1}^{T} \sqrt{ \frac{1}{C} \sum_{c=1}^{C} E_{b,t,c}^{2} }. 
\end{equation} 
\paragraph{RMSE-Element.} RMSE-Element computes RMSE independently for each feature element: 
\begin{equation} 
\mathcal{L}_{\mathrm{Element}} = \frac{1}{BTC} \sum_{b=1}^{B} \sum_{t=1}^{T} \sum_{c=1}^{C} \sqrt{E_{b,t,c}^{2}}. 
\end{equation} 
Since each group contains only one element, RMSE-Element is equivalent to the mean absolute reconstruction error. All variants share the same gradient normalization principle: within each group, the gradient norm depends on the group size rather than the local reconstruction loss magnitude. As the groups are equal-sized within each variant, the overall output-space gradient magnitude remains constant at $1/\sqrt{BTC}$. The variants therefore differ only in the feature grouping over which this normalization is applied.

\section{Full Downstream Evaluation Results}
\label{appendix:full_downstream}

\begin{table*}[th] 
\centering 
\vskip -0.1 in
\caption{Full downstream results under OmniQuant across diverse LLMs and quantization settings. Avg denotes the average accuracy over PIQA, ARC-Easy (ARC-e), ARC-Challenge (ARC-c), Winogrande (WG), HellaSwag (HS), and BoolQ.} 
\label{tab:full_downstream_omniquant} 
\vskip -0.06in 
\renewcommand{\arraystretch}{1.05} 
\setlength{\tabcolsep}{2.5pt} 
\resizebox{0.87\textwidth}{!}{ 
\begin{tabular}{l|c|c|c|cccccc|c} 
\hline Model & \#Bits & Loss & PPL$\downarrow$ & PIQA$\uparrow$ & ARC-e$\uparrow$ & ARC-c$\uparrow$ & WG$\uparrow$ & HS$\uparrow$ & BoolQ$\uparrow$ & Avg$\uparrow$ \\ 
\hline 
\multirow{11}{*}{LLaMA2-7B} & Original & -- & 5.47 & 79.11 & 74.54 & 46.25 & 69.06 & 76.00 & 77.71 & 70.45 \\
\cline{2-11} & \multirow{2}{*}{W4A16} & MSE & 5.75 & 78.35 & 71.63 & 42.83 & 69.53 & 73.90 & 73.73 & 68.33 \\
& & RMSE & 5.70 & 78.02 & 73.57 & 43.43 & 68.35 & 74.46 & 76.57 & 69.07 \\
\cline{2-11} & \multirow{2}{*}{W3A16} & MSE & 6.62 & 75.90 & 65.87 & 38.57 & 65.43 & 69.70 & 68.87 & 64.05 \\
& & RMSE & 6.38 & 76.77 & 69.99 & 42.49 & 67.32 & 72.04 & 72.08 & 66.78 \\
\cline{2-11} & \multirow{2}{*}{W2A16} & MSE & 36.92 & 57.73 & 35.48 & 25.68 & 50.59 & 35.97 & 58.44 & 43.98 \\
& & RMSE & 16.06 & 63.93 & 43.22 & 26.88 & 56.20 & 47.15 & 62.45 & 49.97 \\
\cline{2-11} & \multirow{2}{*}{W2A16G128} & MSE & 11.05 & 66.21 & 46.89 & 27.73 & 56.75 & 51.65 & 60.73 & 51.66 \\
& & RMSE & 9.10 & 70.62 & 59.05 & 31.74 & 59.67 & 57.01 & 65.66 & 57.29 \\
\cline{2-11} & \multirow{2}{*}{W4A4} & MSE & 17.42 & 62.35 & 46.38 & 27.73 & 54.46 & 46.94 & 56.67 & 49.09 \\
& & RMSE & 11.48 & 69.42 & 57.41 & 33.70 & 55.01 & 59.86 & 61.96 & 56.23 \\ 
\hline
\multirow{11}{*}{LLaMA3.1-8B} & Original & -- & 6.24 & 81.18 & 80.89 & 53.67 & 73.56 & 78.87 & 82.08 & 75.04 \\ \cline{2-11} & \multirow{2}{*}{W4A16} & MSE & 7.19 & 80.52 & 78.11 & 50.43 & 71.43 & 76.45 & 78.62 & 72.59 \\
& & RMSE & 6.92 & 80.14 & 78.07 & 50.43 & 71.19 & 77.45 & 80.06 & 72.89 \\
\cline{2-11} & \multirow{2}{*}{W3A16} & MSE & 12.38 & 72.47 & 63.17 & 37.12 & 58.33 & 63.62 & 68.59 & 60.55 \\
& & RMSE & 9.46 & 77.15 & 68.90 & 42.24 & 68.82 & 71.71 & 74.19 & 67.17 \\
\cline{2-11} & \multirow{2}{*}{W2A16} & MSE & 1021.33 & 54.03 & 31.10 & 21.84 & 50.59 & 26.57 & 37.86 & 37.00 \\ 
& & RMSE & 140.63 & 60.01 & 39.86 & 25.09 & 50.83 & 32.15 & 46.73 & 42.44 \\
\cline{2-11} & \multirow{2}{*}{W2A16G128} & MSE & 124.49 & 54.41 & 29.67 & 21.84 & 53.83 & 29.83 & 38.84 & 38.07 \\
& & RMSE & 20.21 & 62.68 & 46.68 & 27.56 & 53.51 & 44.03 & 57.16 & 48.60 \\
\cline{2-11} & \multirow{2}{*}{W4A4} & MSE & 110.88 & 54.46 & 32.07 & 22.87 & 51.46 & 35.33 & 44.34 & 40.09 \\
& & RMSE & 44.29 & 53.32 & 33.84 & 24.57 & 52.72 & 44.11 & 52.48 & 43.51 \\ 
\hline
\multirow{11}{*}{Qwen3.5-9B} & Original & -- & 8.65 & 80.09 & 74.33 & 55.80 & 73.01 & 78.15 & 89.20 & 75.10 \\ 
\cline{2-11} & \multirow{2}{*}{W4A16} & MSE & 9.28 & 78.89 & 76.64 & 54.52 & 71.19 & 76.22 & 87.89 & 74.23 \\
& & RMSE & 9.14 & 80.25 & 75.67 & 55.89 & 71.59 & 76.43 & 86.27 & 74.35 \\
\cline{2-11} & \multirow{2}{*}{W3A16} & MSE & 11.56 & 76.77 & 77.61 & 52.13 & 68.82 & 70.33 & 80.92 & 71.10 \\
& & RMSE & 11.21 & 77.04 & 79.63 & 54.18 & 69.46 & 72.06 & 84.40 & 72.79 \\
\cline{2-11} & \multirow{2}{*}{W2A16} & MSE & 99.05 & 57.18 & 42.47 & 25.77 & 52.96 & 32.23 & 62.02 & 45.44 \\
& & RMSE & 51.56 & 63.77 & 59.43 & 36.86 & 56.75 & 45.64 & 66.24 & 54.78 \\
\cline{2-11} & \multirow{2}{*}{W2A16G128} & MSE & 23.61 & 64.20 & 55.47 & 30.38 & 54.70 & 45.91 & 70.58 & 53.54 \\
& & RMSE & 17.98 & 68.55 & 65.57 & 36.69 & 59.59 & 54.62 & 75.17 & 60.03 \\
\cline{2-11} & \multirow{2}{*}{W4A4} & MSE & 69.74 & 58.05 & 44.87 & 27.90 & 51.93 & 40.62 & 49.24 & 45.43 \\
& & RMSE & 51.88 & 62.84 & 47.10 & 31.14 & 51.85 & 47.61 & 55.66 & 49.37 \\
\hline
\multirow{11}{*}{Qwen3.5-27B} & Original & -- & 6.90 & 82.21 & 79.76 & 61.69 & 78.85 & 83.43 & 76.70 & 77.10 \\
\cline{2-11} & \multirow{2}{*}{W4A16} & MSE & 7.17 & 82.21 & 79.92 & 61.43 & 77.74 & 82.62 & 82.32 & 77.71 \\
& & RMSE & 7.09 & 82.64 & 81.94 & 64.16 & 77.98 & 82.37 & 81.10 & 78.37 \\
\cline{2-11} & \multirow{2}{*}{W3A16} & MSE & 8.42 & 81.39 & 84.55 & 62.63 & 76.01 & 79.77 & 78.38 & 77.12 \\
& & RMSE & 8.26 & 81.61 & 84.68 & 63.57 & 76.64 & 80.20 & 82.11 & 78.13 \\
\cline{2-11} & \multirow{2}{*}{W2A16} & MSE & 37.51 & 68.55 & 71.34 & 44.80 & 60.62 & 53.46 & 63.91 & 60.45 \\
& & RMSE & 22.20 & 74.59 & 78.11 & 52.47 & 68.90 & 67.99 & 83.67 & 70.96 \\
\cline{2-11} & \multirow{2}{*}{W2A16G128} & MSE & 11.73 & 74.76 & 80.39 & 52.82 & 66.93 & 67.65 & 82.72 & 70.88 \\ 
& & RMSE & 11.29 & 75.73 & 81.73 & 55.55 & 70.48 & 70.01 & 85.05 & 73.09 \\
\cline{2-11} & \multirow{2}{*}{W4A4} & MSE & 29.45 & 68.34 & 63.34 & 39.93 & 56.75 & 60.12 & 68.26 & 59.46 \\
& & RMSE & 16.13 & 74.86 & 73.15 & 50.17 & 63.69 & 69.50 & 73.67 & 67.51 \\
\hline 
\multirow{11}{*}{Qwen3.5-35B-A3B} & Original & -- & 6.59 & 83.68 & 79.17 & 62.37 & 74.43 & 82.31 & 88.07 & 78.34 \\
\cline{2-11} & \multirow{2}{*}{W4A16} & MSE & 7.22 & 82.54 & 81.23 & 61.01 & 73.17 & 80.84 & 86.91 & 77.61 \\
& & RMSE & 7.06 & 81.77 & 80.43 & 61.52 & 73.09 & 81.28 & 89.82 & 77.98 \\
\cline{2-11} & \multirow{2}{*}{W3A16} & MSE & 9.87 & 79.76 & 78.62 & 58.62 & 70.24 & 75.03 & 75.99 & 73.05 \\
& & RMSE & 8.83 & 79.43 & 77.23 & 57.59 & 69.93 & 77.39 & 85.41 & 74.50 \\
\cline{2-11} & \multirow{2}{*}{W2A16} & MSE & 73.79 & 61.26 & 42.17 & 26.11 & 49.33 & 37.52 & 64.16 & 46.76 \\
& & RMSE & 21.85 & 69.80 & 56.78 & 37.63 & 56.51 & 56.34 & 68.47 & 57.59 \\
\cline{2-11} & \multirow{2}{*}{W2A16G128} & MSE & 24.66 & 69.10 & 57.53 & 35.49 & 53.59 & 54.84 & 68.26 & 56.47 \\
& & RMSE & 13.90 & 74.81 & 74.07 & 49.15 & 61.09 & 62.02 & 79.42 & 66.76 \\
\cline{2-11} & \multirow{2}{*}{W4A4} & MSE & 12.42 & 72.52 & 68.10 & 44.54 & 59.51 & 71.40 & 77.74 & 65.63 \\
& & RMSE & 10.82 & 75.08 & 66.75 & 46.33 & 63.54 & 74.36 & 78.81 & 67.48 \\ 
\hline
\end{tabular} } 
\vskip -0.00in 
\end{table*}

In the main paper, we report the average downstream accuracy for compactness. In this section, we provide the full per-task results on PIQA, ARC-Easy, ARC-Challenge, Winogrande, HellaSwag, and BoolQ. These results provide a more detailed view of the downstream behavior of MSE and RMSE across different models, quantization settings, and PTQ methods.

\begin{table*}[t!] 
\centering 
\caption{Full downstream results across representative learning-based PTQ methods under W4A4 quantization. Avg denotes the average accuracy over PIQA, ARC-Easy, ARC-Challenge, Winogrande, HellaSwag, and BoolQ.} \label{tab:full_downstream_across_methods} 
\vskip -0.05in 
\renewcommand{\arraystretch}{1.05} 
\setlength{\tabcolsep}{3pt} 
\resizebox{0.84\textwidth}{!}{ 
\begin{tabular}{l|l|c|c|cccccc|c} \hline Model & Method & Loss & PPL$\downarrow$ & PIQA$\uparrow$ & ARC-e$\uparrow$ & ARC-c$\uparrow$ & WG$\uparrow$ & HS$\uparrow$ & BoolQ$\uparrow$ & Avg$\uparrow$ \\
\hline 
\multirow{9}{*}{LLaMA2-7B} & Original & -- & 5.47 & 79.11 & 74.54 & 46.25 & 69.06 & 76.00 & 77.71 & 70.45 \\
\cline{2-11} & \multirow{2}{*}{OmniQuant} & MSE & 17.42 & 62.35 & 46.38 & 27.73 & 54.46 & 46.94 & 56.67 & 49.09 \\
& & RMSE & 11.48 & 69.42 & 57.41 & 33.70 & 55.01 & 59.86 & 61.96 & 56.23 \\
\cline{2-11} & \multirow{2}{*}{SliderQuant} & MSE & 8.48 & 70.46 & 54.92 & 32.85 & 58.09 & 60.60 & 66.30 & 57.20 \\
& & RMSE & 8.14 & 72.09 & 60.69 & 34.98 & 60.77 & 62.81 & 67.09 & 59.74 \\
\cline{2-11} & \multirow{2}{*}{DuQuant} & MSE & 6.07 & 76.71 & 69.11 & 42.49 & 67.88 & 72.01 & 73.24 & 66.91 \\
& & RMSE & 6.01 & 77.31 & 70.62 & 43.77 & 67.01 & 72.72 & 74.71 & 67.69 \\
\cline{2-11} & \multirow{2}{*}{FlatQuant} & MSE & 5.82 & 77.42 & 71.55 & 43.17 & 66.30 & 74.15 & 73.70 & 67.72 \\
& & RMSE & 5.78 & 77.20 & 73.61 & 43.52 & 68.03 & 74.29 & 75.93 & 68.76 \\
\hline 
\multirow{9}{*}{Qwen3.5-9B} & Original & -- & 8.65 & 80.09 & 74.33 & 55.80 & 73.01 & 78.15 & 89.20 & 75.10 \\
\cline{2-11} & \multirow{2}{*}{OmniQuant} & MSE & 69.74 & 58.05 & 44.87 & 27.90 & 51.93 & 40.62 & 49.24 & 45.43 \\
& & RMSE & 51.88 & 62.84 & 47.10 & 31.14 & 51.85 & 47.61 & 55.66 & 49.37 \\
\cline{2-11} & \multirow{2}{*}{SliderQuant} & MSE & 21.41 & 69.64 & 64.39 & 40.44 & 58.96 & 58.64 & 70.49 & 60.43 \\
& & RMSE & 20.64 & 69.21 & 66.58 & 42.24 & 62.19 & 59.94 & 71.35 & 61.92 \\ 
\cline{2-11} & \multirow{2}{*}{DuQuant} & MSE & 10.95 & 76.06 & 73.86 & 47.18 & 66.38 & 71.68 & 83.95 & 69.85 \\
& & RMSE & 10.54 & 77.48 & 73.11 & 50.09 & 67.64 & 72.77 & 82.51 & 70.60 \\
\cline{2-11} & \multirow{2}{*}{FlatQuant} & MSE & 10.26 & 77.75 & 73.65 & 53.50 & 68.59 & 74.92 & 84.95 & 72.23 \\
& & RMSE & 9.15 & 79.16 & 76.56 & 55.38 & 70.88 & 76.15 & 87.00 & 74.19 \\
\hline
\end{tabular} } 
\end{table*}

\section{Additional Results of RMSE Variants}
\label{appendix:rmse_variants}

Tables~\ref{tab:duquant_rmse_variants}, \ref{tab:omni_rmse_llama2}, \ref{tab:omni_rmse_llama31}, \ref{tab:omni_rmse_qwen35_9b}, and \ref{tab:omni_rmse_qwen35_27b} provide additional comparisons of RMSE-Sample, RMSE-Channel, RMSE-Token, and RMSE-Element across learning-based PTQ methods, model families, and quantization settings. All four variants consistently outperform MSE across the evaluated configurations, further supporting the effectiveness of the shared gradient normalization principle across different feature groupings. While the best-performing variant varies across individual configurations, we observe an overall model-family-level preference: RMSE-Token generally performs favorably on LLaMA-family models, whereas RMSE-Channel performs favorably on Qwen-family models. Accordingly, unless otherwise specified, we adopt these variants as the respective defaults in the main experiments. These defaults reflect their overall empirical performance rather than per-configuration optimality. We report broad representative comparisons rather than exhaustively evaluating every possible configuration.

\begin{table*}[!ht] 
\centering 
\caption{Comparison of RMSE variants with DuQuant under W4A4 quantization. Avg denotes the average accuracy over PIQA, ARC-Easy, ARC-Challenge, Winogrande, HellaSwag, and BoolQ.} 
\label{tab:duquant_rmse_variants} 
\vskip -0.05in \renewcommand{\arraystretch}{1.05} 
\setlength{\tabcolsep}{3pt} 
\resizebox{0.88\textwidth}{!}{ \begin{tabular}{l|l|c|cccccc|c} \hline Model & Loss & PPL$\downarrow$ & PIQA$\uparrow$ & ARC-e$\uparrow$ & ARC-c$\uparrow$ & WG$\uparrow$ & HS$\uparrow$ & BoolQ$\uparrow$ & Avg$\uparrow$ \\
\hline 
\multirow{5}{*}{LLaMA2-7B} & MSE & 6.07 & 76.71 & 69.11 & 42.49 & 67.88 & 72.01 & 73.24 & 66.91 \\
& RMSE-Sample & 6.02 & 76.28 & 70.03 & 42.49 & 67.88 & 72.88 & 74.10 & 67.28 \\
& RMSE-Channel & 6.01 & 78.02 & 71.55 & 43.69 & 66.06 & 72.76 & 73.61 & 67.61 \\
& RMSE-Token & 6.01 & 77.31 & 70.62 & 43.77 & 67.01 & 72.72 & 74.71 & 67.69 \\
& RMSE-Element & 6.01 & 77.31 & 70.54 & 42.66 & 65.59 & 72.68 & 73.55 & 67.06 \\
\hline 
\multirow{5}{*}{Qwen3.5-9B} & MSE & 10.95 & 76.06 & 73.86 & 47.18 & 66.38 & 71.68 & 83.95 & 69.85 \\
& RMSE-Sample & 10.41 & 77.20 & 75.13 & 52.65 & 67.32 & 72.56 & 82.05 & 71.15 \\
& RMSE-Channel & 10.54 & 77.48 & 73.11 & 50.09 & 67.64 & 72.77 & 82.51 & 70.60 \\
& RMSE-Token & 10.38 & 77.26 & 74.12 & 51.54 & 66.93 & 72.86 & 78.81 & 70.25 \\
& RMSE-Element & 10.37 & 77.37 & 73.65 & 50.60 & 66.14 & 73.34 & 80.24 & 70.22 \\
\hline 
\end{tabular}} 
\vskip -0.1in 
\end{table*}

\begin{table*}[!ht] 
\centering 
\vskip -0.1 in
\caption{Comparison of RMSE variants with OmniQuant on LLaMA2-7B across quantization settings. Avg denotes the average accuracy over PIQA, ARC-Easy, ARC-Challenge, Winogrande, HellaSwag, and BoolQ.} 
\label{tab:omni_rmse_llama2} 
\vskip -0.05in 
\renewcommand{\arraystretch}{1.05} 
\setlength{\tabcolsep}{3pt} 
\resizebox{0.91\textwidth}{!}{%
\begin{tabular}{c|l|c|cccccc|c} 
\hline 
\#Bits & Loss & PPL$\downarrow$ & PIQA$\uparrow$ & ARC-e$\uparrow$ & ARC-c$\uparrow$ & WG$\uparrow$ & HS$\uparrow$ & BoolQ$\uparrow$ & Avg$\uparrow$ \\
\hline 
\multirow{5}{*}{W4A16} & MSE & 5.75 & 78.35 & 71.63 & 42.83 & 69.53 & 73.90 & 73.73 & 68.33 \\
& RMSE-Sample & 5.70 & 79.05 & 73.23 & 43.52 & 68.51 & 74.19 & 74.77 & 68.88 \\
& RMSE-Channel & 5.70 & 78.40 & 73.53 & 44.11 & 69.46 & 74.33 & 76.15 & 69.33 \\
& RMSE-Token & 5.70 & 78.02 & 73.57 & 43.43 & 68.35 & 74.46 & 76.57 & 69.07 \\
& RMSE-Element & 5.71 & 78.29 & 72.90 & 42.06 & 68.35 & 74.68 & 75.54 & 68.64 \\ 
\hline 
\multirow{5}{*}{W3A16} & MSE & 6.62 & 75.90 & 65.87 & 38.57 & 65.43 & 69.70 & 68.87 & 64.05 \\
& RMSE-Sample & 6.49 & 75.14 & 63.72 & 38.31 & 66.14 & 71.30 & 70.40 & 64.17 \\
& RMSE-Channel & 6.45 & 77.04 & 68.10 & 41.21 & 67.64 & 71.56 & 69.72 & 65.88 \\
& RMSE-Token & 6.38 & 76.77 & 69.99 & 42.49 & 67.32 & 72.04 & 72.08 & 66.78 \\
& RMSE-Element & 6.43 & 77.04 & 68.31 & 41.72 & 67.80 & 72.30 & 72.32 & 66.58 \\
\hline 
\multirow{5}{*}{W2A16} & MSE & 36.92 & 57.73 & 35.48 & 25.68 & 50.59 & 35.97 & 58.44 & 43.98 \\
& RMSE-Sample & 25.58 & 57.73 & 37.50 & 26.19 & 52.80 & 38.19 & 58.13 & 45.09 \\
& RMSE-Channel & 22.46 & 63.00 & 38.59 & 25.68 & 52.57 & 39.70 & 56.51 & 46.01 \\
& RMSE-Token & 16.06 & 63.93 & 43.22 & 26.88 & 56.20 & 47.15 & 62.45 & 49.97 \\
& RMSE-Element & 22.78 & 61.04 & 41.75 & 26.11 & 53.12 & 41.45 & 57.09 & 46.76 \\
\hline 
\multirow{5}{*}{W4A4} & MSE & 17.42 & 62.35 & 46.38 & 27.73 & 54.46 & 46.94 & 56.67 & 49.09 \\
& RMSE-Sample & 14.15 & 63.44 & 47.10 & 29.18 & 54.38 & 53.00 & 63.30 & 51.73 \\
& RMSE-Channel & 13.14 & 67.52 & 52.57 & 31.48 & 53.28 & 57.31 & 58.23 & 53.40 \\
& RMSE-Token & 11.48 & 69.42 & 57.41 & 33.70 & 55.01 & 59.86 & 61.96 & 56.23 \\
& RMSE-Element & 13.64 & 67.90 & 55.13 & 34.39 & 55.41 & 55.58 & 57.68 & 54.35 \\
\hline 
\multirow{5}{*}{W2A16G128} & MSE & 11.05 & 66.21 & 46.89 & 27.73 & 56.75 & 51.65 & 60.73 & 51.66 \\
& RMSE-Sample & 9.20 & 70.02 & 55.64 & 31.23 & 60.14 & 56.61 & 65.84 & 56.58 \\
& RMSE-Channel & 9.13 & 71.33 & 57.24 & 32.68 & 60.30 & 57.13 & 65.50 & 57.36 \\
& RMSE-Token & 9.10 & 70.62 & 59.05 & 31.74 & 59.67 & 57.01 & 65.66 & 57.29 \\ 
& RMSE-Element & 9.22 & 70.89 & 57.58 & 32.17 & 61.72 & 56.26 & 66.61 & 57.54 \\
\hline 
\end{tabular}}
\vspace{1mm} 
\vskip -0.05in 
\end{table*} 

\begin{table*}[!ht] 
\centering 
\caption{Comparison of RMSE variants with OmniQuant on LLaMA3.1-8B across quantization settings. Avg denotes the average accuracy over PIQA, ARC-Easy, ARC-Challenge, Winogrande, HellaSwag, and BoolQ.} 
\label{tab:omni_rmse_llama31} 
\vskip -0.05in 
\renewcommand{\arraystretch}{1.05} \setlength{\tabcolsep}{3pt} 
\resizebox{0.91\textwidth}{!}{%
\begin{tabular}{c|l|c|cccccc|c} \hline \#Bits & Loss & PPL$\downarrow$ & PIQA$\uparrow$ & ARC-e$\uparrow$ & ARC-c$\uparrow$ & WG$\uparrow$ & HS$\uparrow$ & BoolQ$\uparrow$ & Avg$\uparrow$ \\ 
\hline 
\multirow{5}{*}{W4A16} & MSE & 7.19 & 80.52 & 78.11 & 50.43 & 71.43 & 76.45 & 78.62 & 72.59 \\
& RMSE-Sample & 6.96 & 80.90 & 77.95 & 49.66 & 71.74 & 77.17 & 79.20 & 72.77 \\
& RMSE-Channel & 6.96 & 79.92 & 78.45 & 51.11 & 71.67 & 77.34 & 78.75 & 72.87 \\
& RMSE-Token & 6.92 & 80.14 & 78.07 & 50.43 & 71.19 & 77.45 & 80.06 & 72.89 \\
& RMSE-Element & 6.96 & 79.76 & 77.27 & 50.43 & 71.27 & 77.23 & 79.97 & 72.65 \\
\hline \multirow{5}{*}{W3A16} & MSE & 12.38 & 72.47 & 63.17 & 37.12 & 58.33 & 63.62 & 68.59 & 60.55 \\
& RMSE-Sample & 9.88 & 76.82 & 71.42 & 43.26 & 67.48 & 70.16 & 68.87 & 66.34 \\
& RMSE-Channel & 9.56 & 77.42 & 67.34 & 42.83 & 67.56 & 71.29 & 75.66 & 67.02 \\
& RMSE-Token & 9.46 & 77.15 & 68.90 & 42.24 & 68.82 & 71.71 & 74.19 & 67.17 \\
& RMSE-Element & 9.58 & 76.61 & 71.80 & 44.28 & 67.01 & 71.81 & 74.59 & 67.68 \\
\hline 
\multirow{5}{*}{W2A16} & MSE & 1021.33 & 54.03 & 31.10 & 21.84 & 50.59 & 26.57 & 37.86 & 37.00 \\
& RMSE-Sample & 941.96 & 54.46 & 32.24 & 22.27 & 49.33 & 26.58 & 38.07 & 37.16 \\
& RMSE-Channel & 194.77 & 56.47 & 36.57 & 24.74 & 51.14 & 30.76 & 45.17 & 40.81 \\
& RMSE-Token & 140.63 & 60.01 & 39.86 & 25.09 & 50.83 & 32.15 & 46.73 & 42.44 \\
& RMSE-Element & 368.30 & 55.98 & 34.05 & 22.35 & 50.83 & 27.15 & 38.41 & 38.13 \\
\hline 
\multirow{5}{*}{W4A4} & MSE & 110.88 & 54.46 & 32.07 & 22.87 & 51.46 & 35.33 & 44.34 & 40.09 \\
& RMSE-Sample & 68.36 & 55.50 & 32.32 & 22.27 & 48.54 & 39.49 & 50.95 & 41.51 \\
& RMSE-Channel & 55.16 & 53.65 & 30.35 & 24.15 & 52.64 & 43.85 & 49.05 & 42.28 \\
& RMSE-Token & 44.29 & 53.32 & 33.84 & 24.57 & 52.72 & 44.11 & 52.48 & 43.51 \\
& RMSE-Element & 59.20 & 57.07 & 34.68 & 23.21 & 52.64 & 37.88 & 52.17 & 42.94 \\
\hline 
\multirow{5}{*}{W2A16G128} & MSE & 124.49 & 54.41 & 29.67 & 21.84 & 53.83 & 29.83 & 38.84 & 38.07 \\
& RMSE-Sample & 20.52 & 63.11 & 45.79 & 26.45 & 52.72 & 43.92 & 61.74 & 48.96 \\
& RMSE-Channel & 20.91 & 64.74 & 47.14 & 27.05 & 55.41 & 45.55 & 58.20 & 49.68 \\
& RMSE-Token & 20.21 & 62.68 & 46.68 & 27.56 & 53.51 & 44.03 & 57.16 & 48.60 \\
& RMSE-Element & 23.72 & 61.26 & 45.03 & 26.02 & 53.12 & 43.86 & 55.20 & 47.42 \\
\hline 
\end{tabular}} 
\vskip 
-0.05in 
\end{table*}

\begin{table*}[t!] 
\centering 
\caption{Comparison of RMSE variants with OmniQuant on Qwen3.5-9B across quantization settings. Avg denotes the average accuracy over PIQA, ARC-Easy, ARC-Challenge, Winogrande, HellaSwag, and BoolQ.} 
\label{tab:omni_rmse_qwen35_9b} 
\vskip -0.05in 
\renewcommand{\arraystretch}{1.05} 
\setlength{\tabcolsep}{3pt} 
\resizebox{0.91\textwidth}{!}{%
\begin{tabular}{c|l|c|cccccc|c} 
\hline \#Bits & Loss & PPL$\downarrow$ & PIQA$\uparrow$ & ARC-e$\uparrow$ & ARC-c$\uparrow$ & WG$\uparrow$ & HS$\uparrow$ & BoolQ$\uparrow$ & Avg$\uparrow$ \\
\hline \multirow{5}{*}{W4A16} & MSE & 9.28 & 78.89 & 76.64 & 54.52 & 71.19 & 76.22 & 87.89 & 74.23 \\
& RMSE-Sample & 9.14 & 79.82 & 78.62 & 56.23 & 70.48 & 76.46 & 88.20 & 74.97 \\
& RMSE-Channel & 9.14 & 80.25 & 75.67 & 55.89 & 71.59 & 76.43 & 86.27 & 74.35 \\
& RMSE-Token & 9.20 & 79.16 & 78.87 & 58.11 & 71.03 & 76.49 & 86.79 & 75.08 \\
& RMSE-Element & 9.20 & 79.05 & 77.40 & 56.57 & 70.88 & 76.45 & 87.83 & 74.70 \\
\hline 
\multirow{5}{*}{W3A16} & MSE & 11.56 & 76.77 & 77.61 & 52.13 & 68.82 & 70.33 & 80.92 & 71.10 \\
& RMSE-Sample & 11.32 & 77.75 & 80.01 & 53.84 & 68.82 & 71.62 & 87.00 & 73.17 \\
& RMSE-Channel & 11.21 & 77.04 & 79.63 & 54.18 & 69.46 & 72.06 & 84.40 & 72.79 \\
& RMSE-Token & 11.16 & 77.69 & 77.19 & 52.65 & 69.06 & 71.51 & 86.73 & 72.47 \\
& RMSE-Element & 11.29 & 77.58 & 77.82 & 53.24 & 70.24 & 72.11 & 86.54 & 72.92 \\
\hline 
\multirow{5}{*}{W2A16} & MSE & 99.05 & 57.18 & 42.47 & 25.77 & 52.96 & 32.23 & 62.02 & 45.44 \\
& RMSE-Sample & 82.46 & 58.32 & 51.52 & 26.96 & 52.57 & 34.12 & 62.17 & 47.61 \\
& RMSE-Channel & 51.56 & 63.77 & 59.43 & 36.86 & 56.75 & 45.64 & 66.24 & 54.78 \\
& RMSE-Token & 88.41 & 58.16 & 49.62 & 28.41 & 52.25 & 34.37 & 62.97 & 47.63 \\
& RMSE-Element & 61.76 & 62.24 & 56.48 & 33.11 & 57.62 & 43.24 & 63.85 & 52.76 \\
\hline 
\multirow{5}{*}{W4A4} & MSE & 69.74 & 58.05 & 44.87 & 27.90 & 51.93 & 40.62 & 49.24 & 45.43 \\ 
& RMSE-Sample & 66.52 & 59.47 & 48.91 & 30.97 & 50.51 & 42.26 & 54.98 & 47.85 \\
& RMSE-Channel & 51.88 & 62.84 & 47.10 & 31.14 & 51.85 & 47.61 & 55.66 & 49.37 \\
& RMSE-Token & 62.15 & 59.14 & 46.34 & 29.44 & 51.22 & 42.04 & 56.12 & 47.38 \\
& RMSE-Element & 54.12 & 61.48 & 46.59 & 29.44 & 51.38 & 46.66 & 57.19 & 48.79 \\
\hline 
\multirow{5}{*}{W2A16G128} & MSE & 23.61 & 64.20 & 55.47 & 30.38 & 54.70 & 45.91 & 70.58 & 53.54 \\
& RMSE-Sample & 17.85 & 67.36 & 67.05 & 37.46 & 61.01 & 53.91 & 79.69 & 61.08 \\
& RMSE-Channel & 17.98 & 68.55 & 65.57 & 36.69 & 59.59 & 54.62 & 75.17 & 60.03 \\
& RMSE-Token & 18.06 & 67.74 & 65.66 & 37.88 & 61.17 & 53.87 & 73.00 & 59.89 \\ 
& RMSE-Element & 18.68 & 68.50 & 61.62 & 35.92 & 58.80 & 53.17 & 73.76 & 58.63 \\
\hline 
\end{tabular}} 
\vskip -0.0in 
\end{table*}

\begin{table*}[t!] 
\centering 
\caption{Comparison of RMSE variants with OmniQuant on Qwen3.5-27B under W2A16 quantization. Avg denotes the average accuracy over PIQA, ARC-Easy, ARC-Challenge, Winogrande, HellaSwag, and BoolQ.} 
\label{tab:omni_rmse_qwen35_27b} 
\vskip -0.05in 
\renewcommand{\arraystretch}{1.05} 
\setlength{\tabcolsep}{3pt} 
\resizebox{0.88\textwidth}{!}{%
\begin{tabular}{c|l|c|cccccc|c} 
\hline \#Bits & Loss & PPL$\downarrow$ & PIQA$\uparrow$ & ARC-e$\uparrow$ & ARC-c$\uparrow$ & WG$\uparrow$ & HS$\uparrow$ & BoolQ$\uparrow$ & Avg$\uparrow$ \\
\hline \multirow{5}{*}{W2A16} & MSE & 37.51 & 68.55 & 71.34 & 44.80 & 60.62 & 53.46 & 63.91 & 60.45 \\
& RMSE-Sample & 24.28 & 72.96 & 77.90 & 50.60 & 66.46 & 61.73 & 69.36 & 66.50 \\
& RMSE-Channel & 22.20 & 74.59 & 78.11 & 52.47 & 68.90 & 67.99 & 83.67 & 70.96 \\
& RMSE-Token & 25.04 & 72.58 & 76.14 & 47.70 & 65.98 & 59.64 & 72.08 & 65.69 \\
& RMSE-Element & 24.53 & 72.85 & 74.62 & 49.83 & 68.11 & 64.07 & 84.28 & 68.96 \\
\hline 
\end{tabular}} 
\vskip -0.1in 
\end{table*}

\section{Additional Analysis of Optimization Dynamics} 
\label{appendix:optimization_statistics} 
\begin{figure*}[ht!] 
\centering 
\vskip -0.05in \centerline{ \includegraphics[width=0.99\textwidth]{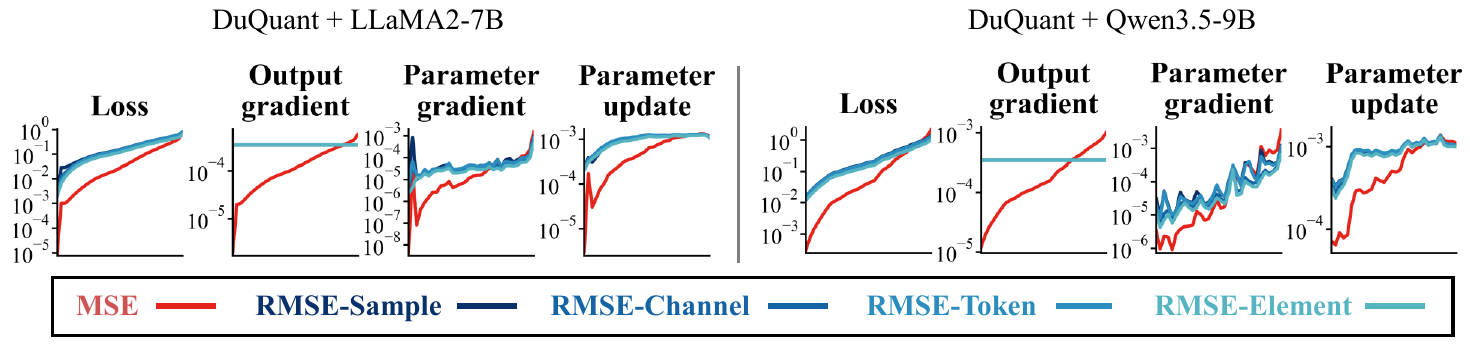} } 
\vskip -0.1in 
\caption{Optimization statistics under MSE and different RMSE variants with DuQuant on LLaMA2-7B and Qwen3.5-9B under W4A4 quantization. The horizontal axis denotes the quantization stages from early to late. } \label{fig:duquant_rmse_variants} 
\vskip -0.15in 
\end{figure*}

\begin{figure*}[t!] 
\centering 
\vskip -0.05in 
\centerline{ \includegraphics[width=0.99\textwidth]{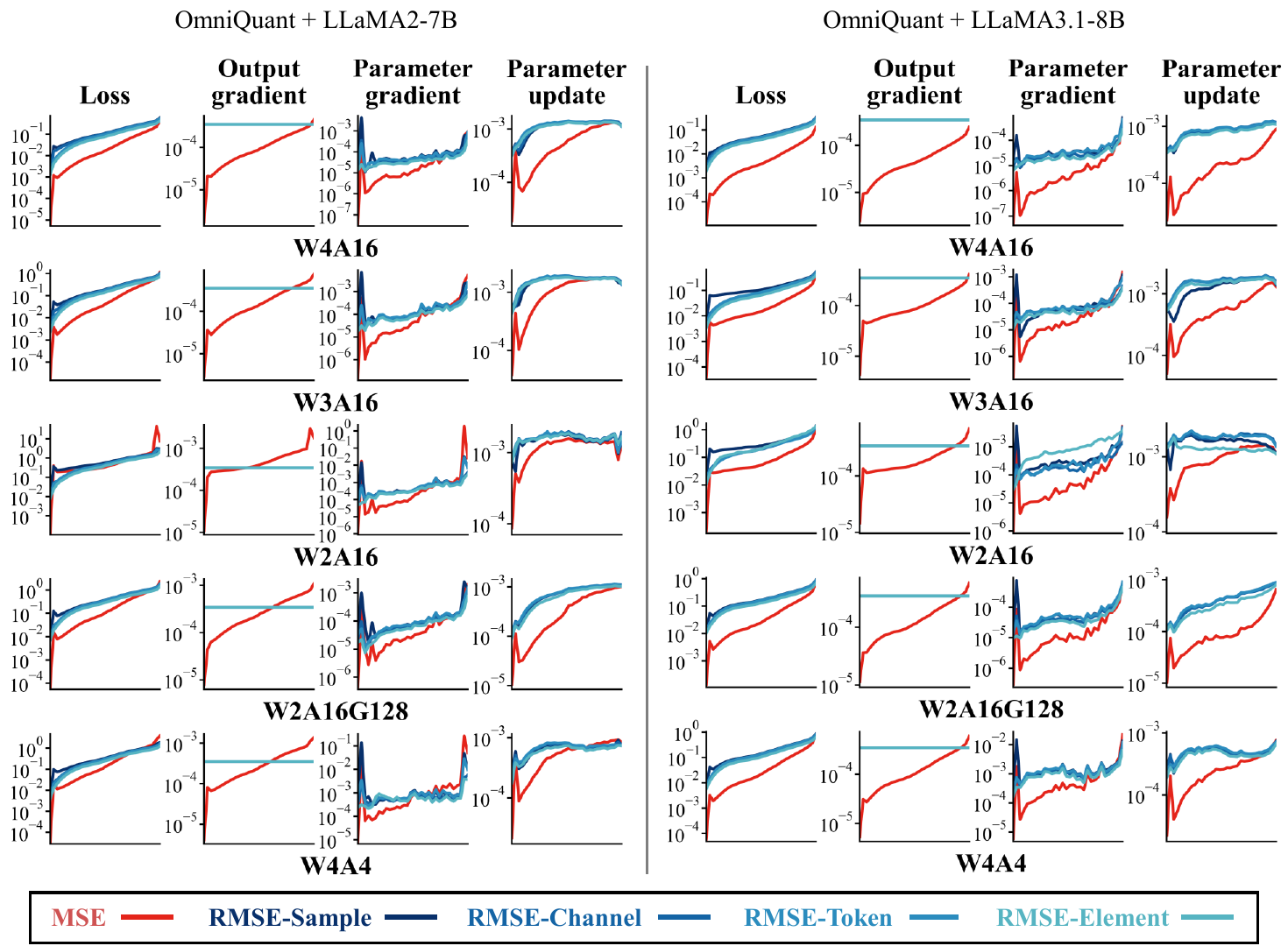} } 
\vskip -0.1in 
\caption{ Optimization statistics under MSE and different RMSE variants with OmniQuant on LLaMA2-7B and LLaMA3.1-8B across quantization settings. The horizontal axis denotes the quantization stages from early to late. } 
\label{fig:omniquant_llama_rmse_variants} 
\vskip -0.15in 
\end{figure*}

\begin{figure*}[t!] 
\centering 
\vskip -0.05in \centerline{ \includegraphics[width=0.99\textwidth]{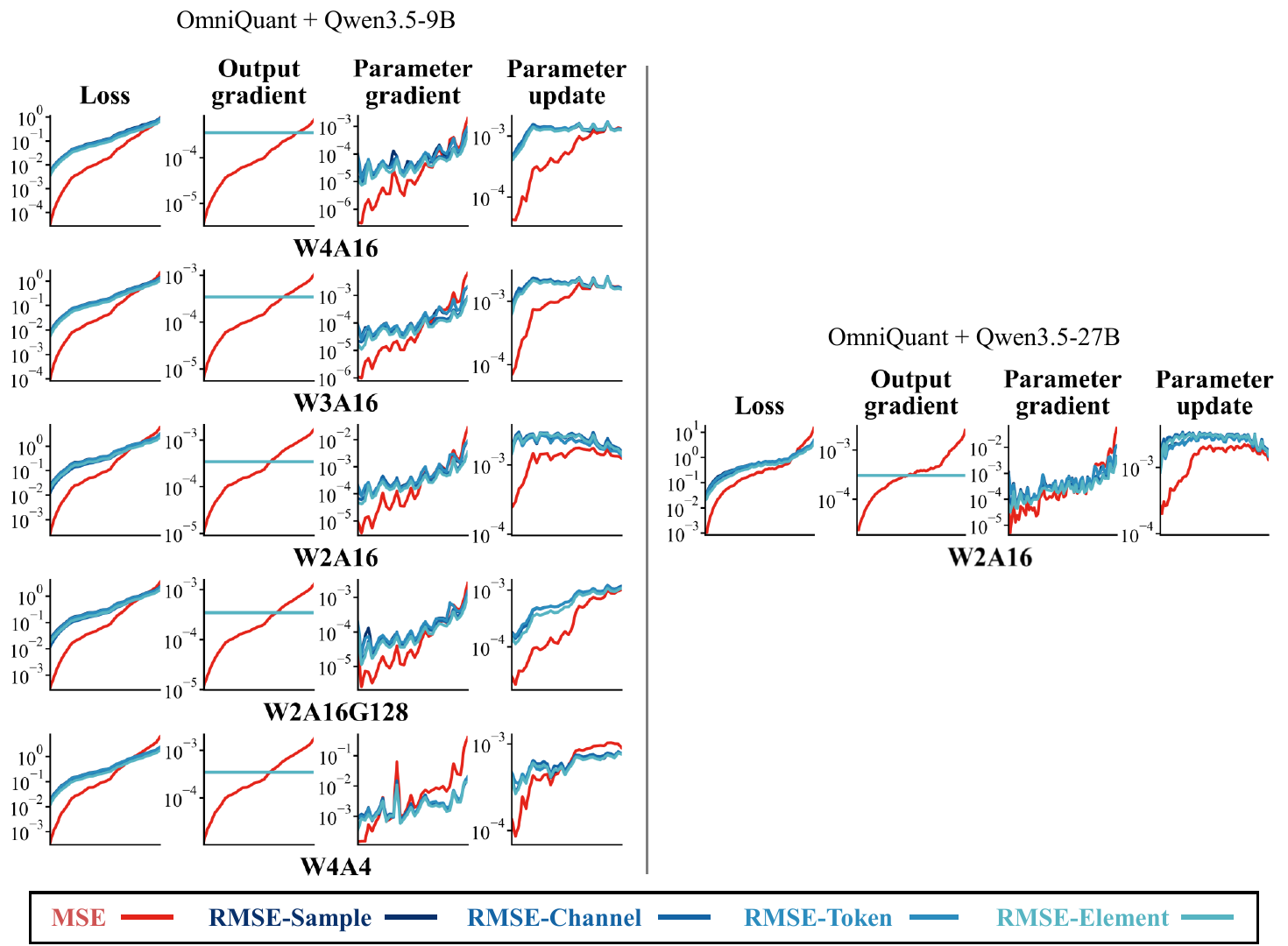} } 
\vskip -0.1in 
\caption{Optimization statistics under MSE and different RMSE variants with OmniQuant. Results are reported across multiple quantization settings on Qwen3.5-9B and under W2A16 quantization on Qwen3.5-27B. The horizontal axis denotes the quantization stages from early to late.}
\label{fig:omniquant_qwen35_rmse_variants} 
\vskip -0.15in 
\end{figure*}

Complementing the optimization statistics presented in Figs.~\ref{fig:across_models_and_settings} and~\ref{fig:across_methods}, Figs.~\ref{fig:duquant_rmse_variants}, \ref{fig:omniquant_llama_rmse_variants}, and \ref{fig:omniquant_qwen35_rmse_variants} provide additional representative results across learning-based PTQ methods, model families, and quantization settings. These figures further compare all four RMSE variants, demonstrating consistent balanced optimization behavior across different feature groupings.

\clearpage
\begin{figure*}[t!] 
\centering 
\vskip -0.15 in
\includegraphics[width=0.99\textwidth]{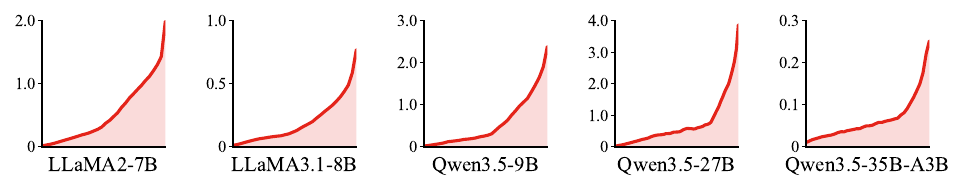} 
\vskip -0.15 in
\caption{Mean absolute magnitudes of high-precision block outputs across model depth. The horizontal axis denotes the block order from early to late, and the vertical axis reports the mean absolute output magnitude, $\|X_{i+1}\|_1/N$, of each block. Statistics are collected from the original high-precision models using 128 WikiText2 samples with a sequence length of 2048.} 
\label{fig:activation_magnitude} 
\vskip -0.1 in
\end{figure*}

\section{Output Magnitudes Across Blocks}
As discussed in Sec.~\ref{sec:optimization_imbalance}, reconstruction losses are computed directly over unit outputs, making their magnitudes inherently sensitive to the magnitudes of the reconstructed features. To examine this factor directly, we measure the mean absolute magnitude of the high-precision block outputs, 
\begin{equation} 
\frac{1}{N}\left\|X_{i+1}\right\|_1, 
\end{equation} 
using 128 WikiText2 calibration samples. As shown in Fig.~\ref{fig:activation_magnitude}, output magnitudes vary substantially across blocks and generally increase with model depth across all evaluated model families, scales, and architectures. Since these statistics are collected from high-precision models without quantization, the observed variation is intrinsic to the model outputs rather than introduced by the quantization process. This cross-block variation provides a natural source of the broad reconstruction loss scale observed during sequential quantization.

\begin{figure*}[t!] 
\centering 
\includegraphics[width=0.9\textwidth]{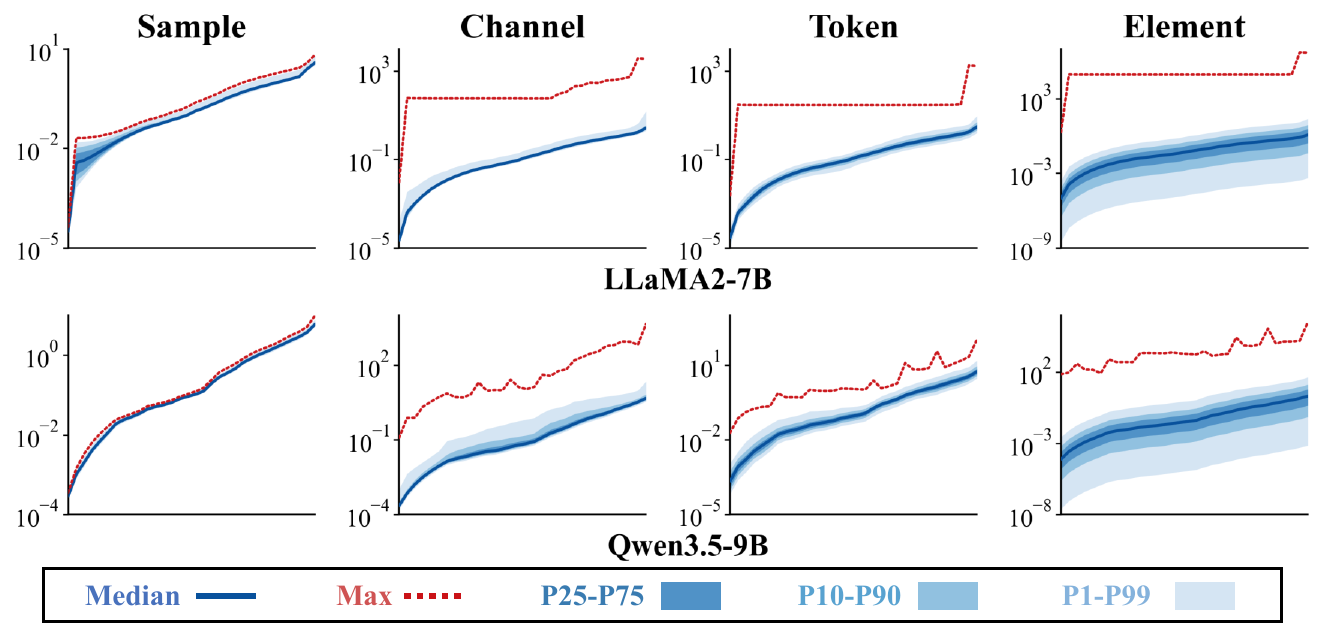} 
\vskip -0.1 in
\caption{Distributions of group-wise reconstruction losses across blocks. For the fully quantized model, we partition the high-precision and quantized outputs of each block into groups at the sample, channel, token, and element levels, and compute the MSE loss independently within each group. The horizontal axis denotes the block order from early to late, and the vertical axis reports the group-wise MSE. The solid line denotes the median across groups, the shaded regions show different percentile intervals, and the red dotted line denotes the maximum. Results are collected from OmniQuant models trained with MSE under W4A4 quantization.}
\label{fig:groupwise_loss_distribution} 
\vskip -0.1 in
\end{figure*}

\section{Reconstruction Loss Magnitude Variation Within Stages} 
To examine whether reconstruction loss magnitudes also vary within each unit, we analyze OmniQuant models trained with MSE and compute the group-wise MSE between the high-precision and quantized unit outputs using the four feature groupings introduced in Sec.~\ref{sec:rmse_family}. As shown in Fig.~\ref{fig:groupwise_loss_distribution}, reconstruction loss magnitudes vary substantially across samples, with increasingly dispersed distributions under the channel, token, and element groupings. This within-unit variation further demonstrates that feature grouping is a meaningful design dimension for applying gradient normalization within each unit.

\end{document}